\def\paperID{9124}
\def\confName{CVPR}
\def\confYear{2024}

\def\paperTitle{Inpaint4DNeRF: Promptable Spatio-Temporal NeRF Inpainting with \\ Generative
Diffusion Models
}

\def\authorBlock{
    Han Jiang\thanks{Equal contribution. The order of authorship was determined alphabetically.} \\
      HKUST\\
      \tt\small hjiangav@connect.ust.hk \and
      Haosen Sun\footnotemark[1] \\
      HKUST\\
      \tt\small hsunas@connect.ust.hk \and
      Ruoxuan Li\footnotemark[1] \\
      HKUST\\
      \tt\small rliba@connect.ust.hk \and
      Chi-Keung Tang \\
      HKUST\\
      \tt\small cktang@cs.ust.hk \and
      Yu-Wing Tai \\
      Dartmouth College\\
      \tt\small yu-wing.tai@dartmouth.edu
}

\newif\ifreview 
\newif\ifarxiv \newcommand{\arxiv}{\arxivtrue}
\newif\ifcamera 
\newif\ifrebuttal 

\arxiv 

\pdfoutput=1
\documentclass[10pt,twocolumn,letterpaper]{article}
\ifreview \usepackage[review]{cvpr} \fi
\ifarxiv \usepackage[pagenumbers]{cvpr} \fi
\ifrebuttal \usepackage[rebuttal]{cvpr} \fi
\ifcamera \usepackage{cvpr} \fi

\usepackage{graphicx}	
\usepackage{amsmath}	
\usepackage{amssymb}	
\usepackage{booktabs}
\usepackage{times}
\usepackage{microtype}
\usepackage{epsfig}
\usepackage[table,xcdraw,dvipsnames]{xcolor}
\usepackage{caption}
\usepackage{float}
\usepackage{placeins}
\usepackage{color, colortbl}
\usepackage{stfloats}
\usepackage{enumitem}
\usepackage{tabularx}
\usepackage{xstring}
\usepackage{multirow}
\usepackage{xspace}
\usepackage{url}
\usepackage{subcaption}
\usepackage{xcolor}
\usepackage[hang,flushmargin]{footmisc}
\usepackage{comment}

\ifcamera \usepackage[accsupp]{axessibility} \fi

% ...

\ifarxiv  \fi

\newcommand{\R}[1]{{%
    \textbf{%
        \ifstrequal{#1}{1}{\textcolor{red}{R#1}}{%
        \ifstrequal{#1}{2}{\textcolor{blue}{R#1}}{%
        \ifstrequal{#1}{3}{\textcolor{magenta}{R#1}}{%
        \ifstrequal{#1}{4}{\textcolor{teal}{R#1}}{%
                           \textcolor{cyan}{R#1}%
        }}}}%
    }%
}}

\usepackage{xr-hyper}

\makeatletter
\newcommand*{\addFileDependency}[1]{
  \typeout{(#1)}
  \@addtofilelist{#1}
  \IfFileExists{#1}{}{\typeout{No file #1.}}
}

\makeatother

\definecolor{cvprblue}{rgb}{0.21,0.49,0.74}
\usepackage[pagebackref,breaklinks,colorlinks,citecolor=cvprblue]{hyperref}
\usepackage[capitalize]{cleveref}
\crefname{section}{Sec.}{Secs.}
\crefname{table}{Table}{Tables}
\crefname{figure}{Fig.}{Figs.}

\begin{document}
\title{\paperTitle}
\author{\authorBlock}
\maketitle

\begin{abstract}
Current Neural Radiance Fields (NeRF) can generate photorealistic novel
views. For editing 3D scenes represented by NeRF, 
with the advent of generative
models,  
this paper proposes Inpaint4DNeRF to
capitalize on state-of-the-art stable diffusion models (e.g.,
ControlNet~\cite{controlnet}) for \textbf{direct} generation of the
underlying completed background content, regardless of static or dynamic. The
key advantages of this generative approach for NeRF inpainting are
twofold. 
First, after rough mask propagation, to complete or fill in previously occluded content, we can
individually generate a small subset of completed images with plausible
content, called seed images, from which simple 3D geometry proxies can be
derived.  Second and the remaining problem is thus 3D multiview
consistency among all completed images, now guided by the seed images and
their 3D proxies.  Without other bells and whistles, our generative Inpaint4DNeRF baseline framework is general which
can be readily extended to 4D dynamic NeRFs, where temporal consistency
can be naturally handled in a similar way as our multiview consistency.
\end{abstract}
\section{Introduction}
\label{sec:intro}
Recent development of Neural Radiance Fields (NeRF)~\cite{mildenhall2020nerf} and its dynamic variants including~\cite{nerfplayer, park2021hypernerf, kplanes_2023} have shown their great potential in modeling scenes in 3D and 4D. This representation is suitable for scene editing from straightforward user inputs such as text prompts, eliminating the need for modeling and animating in detail. One important task in scene editing is \textit{generative inpainting}, which refers to generating plausible content that is consistent with the background scene. Generative inpainting has a wide range of potential applications, including digital art creation, and VR/AR.

Although several recent works have addressed the inpainting and text-guided content generation problem on NeRFs, they have various limitations in directly generating consistent novel content seamlessly with the background based on the text input. Specifically, ~\cite{haque2023instruct,shao2023control4d,zhang2023dyne} allow users to edit the appearance of an existing object in the NeRF based on the text prompt, but their generations are limited by the original target object, thus they cannot handle substantial geometry changes. In~\cite{inpaintnerf360, spinnerf}  a given NeRF is inpainted by removing the target object and inferring the background, but their inpainting task is not generative as they do not match the inferred background with any other user input. Recent generative works~\cite{lin2023magic3d,poole2022dreamfusion,singer2023text4d,wang2023prolificdreamer} produce new 3D or 4D content from the text input, but the generation is not conditioned on the existing background. While~\cite{song2023blendingnerf} can generate contents around the target object without removing it, 
generative inpainting on NeRF should allow users to remove target objects while filling the exposed region with plausible 3D content, which was previously totally or partially occluded in the given scene. 


While applying recent diffusion models~\cite{Rombach_2022_CVPR} on generative 2D inpainting may largely solve the problem, 
simply extending diffusion models to higher dimensions to static or dynamic NeRFs introduces extensive challenges, including new and more complex network structures and sufficient higher-dimensional training data. On the other hand, NeRFs are continuous multiview estimations on its training images which provide a natural connection between 2D images and NeRF, enabling us to propagate the 2D inpainting results by diffusion models to the underlying scene.

Thus, in this paper, we propose Inpaint4DNeRF, the first work on text-guided generative NeRF inpainting with diffusion models, which can be naturally extended to inpaint 4D dynamic NeRFs. Given a text prompt, and a target foreground region specified by the user by text or on the single individual image(s), our method first utilizes stable diffusion to inpaint a few 
independent training views. Next, we regard these inpainted views as seed images and infer a coarse geometry proxy from them. With the strong guidance by seed images and their geometry proxies, the other views being inpainted are constrained to be consistent with the seed images when refined with stable diffusion, which is followed by NeRF finetuning with progressive updates on the training views to obtain final multiview convergence. For dynamic cases, once generative inpainting is achieved on a static single frame, it can be regarded as the seed frame where the edited information can be propagated to other frames to naturally inpaint the scene in 4D without bells and whistles. Our generative Inpaint4DNeRF proposes a text-guided approach for
 inpainting the relevant background after removing static
foreground objects specified by a user-supplied prompt, where multiview consistency is meticulously maintained.

To summarize,  Inpaint4DNeRF presents the following contributions. First, harnessing the power of recent diffusion models on image inpainting, we can directly generate text-guided content with new geometry while being consistent with the context given by the unmasked background. Second, our approach infers and refines other views from initially inpainted seed images to achieve multiview consistency across all the given views, where temporal consistency is naturally enforced and achieved. 

\section{Related Work}
\label{sec:related}

\subsection{NeRF Editing}
NeRF Editing, as pioneered by \cite{zhang2021stnerf,yuan2022nerf,liu2021editing}, capitalizes on the capabilities of Neural Radiance Fields (NeRFs) to enable intricate manipulations and edits within 3D scenes and objects, including object removal, geometry transformations, and appearance editing. Enabling editing within the implicit density and radiance field has enormous promise across many fields, including virtual reality, augmented reality, content development, and beyond. Nevertheless, these prior works either edit the geometry of already existing contents in NeRF, or edit only the appearance of the scene. None of these works targets at \textit{generating} novel contents consistent with the underlying NeRF given by the input images to be inpainted. For this task, ~\cite{zhang2023dyne} generates new texture on an existing object, but does not generate new geometry;~\cite{song2023blendingnerf} generates promptable geometrical content around a target object, but the generation is limited near the object and cannot replace the existing object. In our work, we focus on a subtask of NeRF editing, \textit{generative inpainting} in NeRF. 

\subsection{Inpainting Techniques}
The volume-renderable representation and the training mode of NeRF introduce a natural bridge between the represented 3D or 4D scene and its multiview observations. This enables image editing techniques to be applied on NeRF inpainting.

Recent advancements in 2D image inpainting have primarily relied on generative models \cite{suvorov2021resolution}, particularly Generative Adversarial Networks (GANs) \cite{yu2018generative,yu2018free} and Stable Diffusion (SD) models \cite{repaint, rombach2022highresolution}. These models have demonstrated remarkable capabilities in generating visually plausible predictions for missing pixels.
Specifically, the capacity to simulate complex data distributions has drawn notable attention to Stable Diffusion~\cite{rombach2022highresolution}, an extension of the denoising diffusion probabilistic model that supports inpainting guided by text prompts. This method produces inpainted pixels and samples of superior quality by using controlled diffusion processes. Diffusion models and noise-conditioned score networks are vital to stable diffusion as they interact to optimize the generation process. Researchers have leveraged Stable Diffusion inpainting for various applications, including semantic inpainting, texture synthesis, and realistic object removal. These applications involve understanding contextual information in the image and generating inpaintings that seamlessly integrate with the existing scene.

However, most of these works have been limited to synthesis in the image domain and have yet to explicitly consider maintaining fidelity to the 3D structure across various viewpoints. Several prior works~\cite{inpaintnerf360, spinnerf} addressed the pure NeRF inpainting problem, which removes target objects and infers the background without generating prompt-guided contents. Notably, our study not only goes beyond traditional 2D image inpainting, addressing the challenge of inpainting within 3D and 4D scenes while maintaining view consistency even under perspective changes, but is also promptable in the sense that the inpainted content is matched with the text description while consistent with the underlying 3D or 4D spatiotemporal scene to be inpainted.

\subsection{Text-Guided Visual Content Generation} 
The advent of generative models has led to extensive research on guiding the generation results using natural language. For example, the latent diffusion model, as exemplified by \cite{rombach2022highresolution}, has made significant strides in text-guided image generation. Various image modification techniques, such as 
\cite{li2020manigan, kim2022diffusionclip, li2020lightweight, xia2021tedigan}, have emerged as a result of these improvements.

Based on the above text-to-image achievements, text-to-3D generation has been introduced, as shown in \cite{poole2022dreamfusion, wang2023prolificdreamer, lin2023magic3d, Chen_2023_ICCV}. These approaches aim to bridge the gap in 3D content generation, leveraging the Score Distillation Sampling (SDS) technique and its variants for multiview convergence. Moreover, attempts have been made to generate 4D dynamic content from text~\cite{singer2023text4d}, with several techniques including a temporal consistency regularizer to extend DreamFusion~\cite{poole2022dreamfusion} to dynamic NeRFs. Despite the complexity of SDS sampling, they have achieved impressive results in terms of 3D consistency. On the other hand, our proposed method conditions on the \textit{seed} view to control the generation of other views to force multiview convergence. This approach restricts the ill-posed 
text-guided generation problem to a well-posed problem with strong priors, thus making the problem easier to tackle. 3D generation conditioned on one generated view has been presented in some most recent works, including Zero123~\cite{liu2023zero1to3} and SyncDreamer~\cite{liu2023syncdreamer}. Given an image of an object and multiple camera poses, they can infer feasible observation of the object from other views.

However, existing implementations are all limited to a single object without conditioning on the background, and struggle with manipulating objects within large scenes. In other words, they all contribute to the pure generation task which is different from our inpainting task. Our method distinguishes itself by enabling the removal, addition, and manipulation of specific objects within a given background NeRF, while maintaining consistency with the unmasked background and partially masked foreground objects. In addition, the 3D inpainted results can be extended to 4D while maintaining temporal consistency.

\section{Method}
\begin{figure*}[tp]
    \centering
    \includegraphics[width=\linewidth]{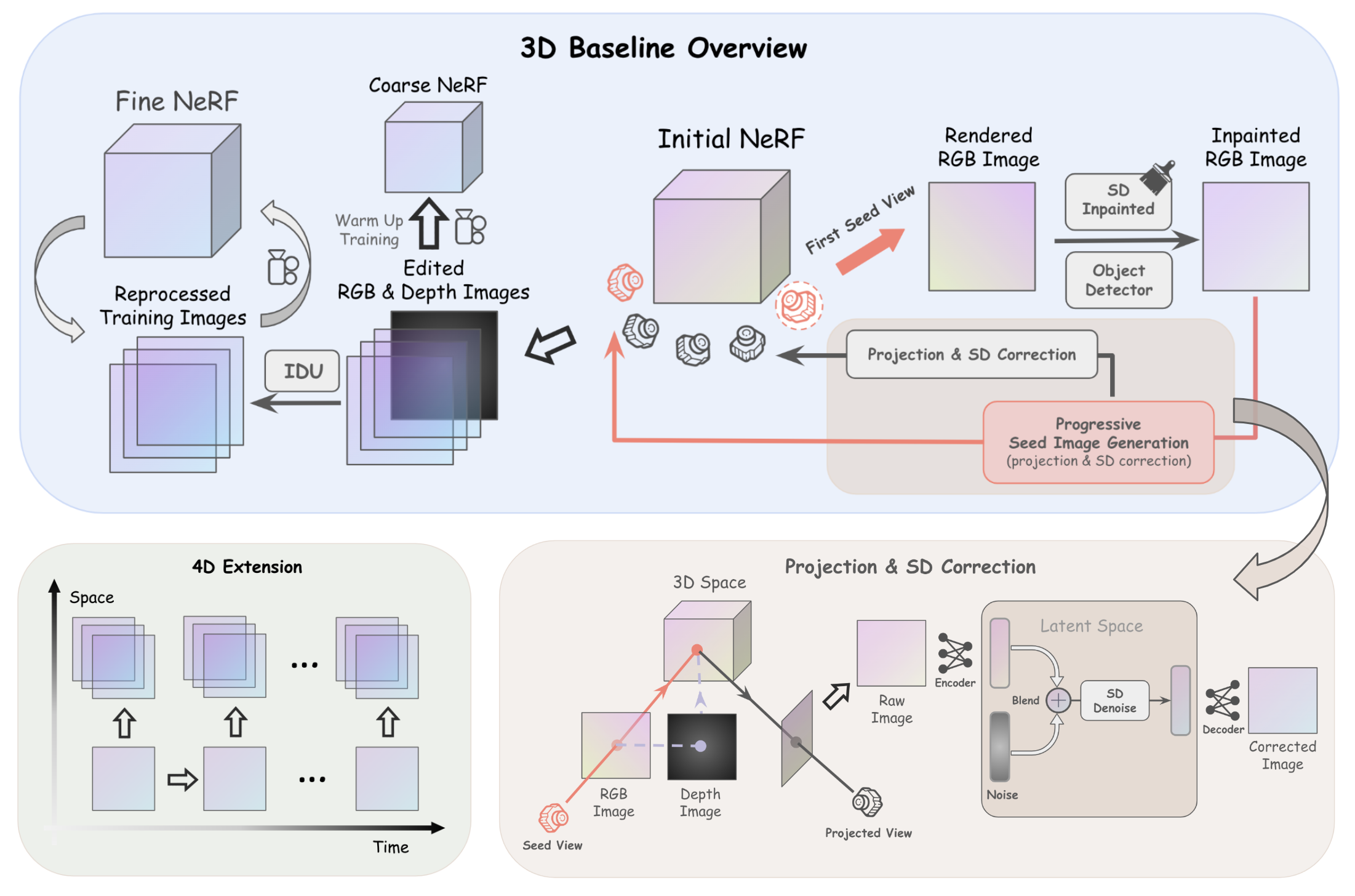}
    \caption{Baseline Overview. Our generative NeRF inpainting is based on the inpainted image of one training view. The other seed images and training images are obtained by using stable diffusion to hallucinate the corrupted detials of the unproject-projected raw image. These images are then used to finetune the NeRF, with warmup training to get geometric and  coarse appearance convergence, followed by iterative training image update to get fine convergence. For 4D extension, we first obtain a temporally consistent inpainted seed video based on the first seed image. Then for each frame, we infer inpainted images on other views by projection and correction, as in our 3D baseline.}
    \label{fig:baseline}
\end{figure*}
\label{sec:method}

Given a pre-trained NeRF, a set of masks on its training images denoting the target object to be replaced (or removed), and a text prompt, we propose \textit{generative promptable inpainting}, which can be decomposed into three objectives: 1) 3D and 4D visual content generation, where the resulting fine-tuned NeRF should contain a new object that is multiview and temporally consistent; 2) text-prompt guided generation, where the semantics of the generated object should match the input text prompt; 3) the generated inpainted content should be consistent with the existing NeRF background.

Our proposed framework consists of three main stages, as shown in Figure~\ref{fig:baseline}.
First, we employ stable diffusion~\cite{Rombach_2022_CVPR} to inpaint one view as the first seed image, and generate a coarse set of seed images conditioned on the first seed image. The other views are then inferred from the seed images and refined by stable diffusion. This stage pre-processes the training images aiming to make convergence easier and faster later. Next, we fine-tune the  NeRF by performing a stable diffusion version of iterative dataset update~\cite{haque2023instruct} to enforce 3D multiview consistency. A converged 3D NeRF is obtained in this stage. If we target at inpainting 4D NeRF, we propagate the 3D inpainted result along the time dimension in the final stage.
In the following, we will describe the three stages in detail, respectively in each subsection.

\subsection{Training View Pre-processing}

Text-guided visual content generation is inherently a highly underdetermined problem: for a given text prompt, there are infinitely many object appearances that are feasible matches. In our method, we generate content in NeRF by first performing inpainting on its training images and then backward propagating the modified pixels of the images into NeRF. 

If we inpaint each view independently, the inpainted content will not be consistent across multiple views. Some prior 3D generation works, using techniques including score distillation sampling~\cite{poole2022dreamfusion}, have demonstrated the possibility of multiview convergence based on independently modified training views. Constraining the generation problem by enforcing the training views to be strongly related to each other will simplify convergence to a large extent. Therefore, we first inpaint a small number of \textit{seed images} associated with a coarse set of cameras that covers sufficiently wide viewing angles. For the other views, the inpainted content are strongly conditioned on these seed images.

\vspace{0.5mm}
\noindent \textbf{Progressive seed image generation.}\quad Given a set of training viewpoints $\{ \mathbf{\phi}^{i} \}$, we manually select a subset $\{ \mathbf{\phi}^{i}_{seed} \} \subseteq \{ \mathbf{\phi}^{i} \}$ that has comparable view coverage with $\{ \mathbf{\phi}^{i} \}$, and perform inpainting on their associated training images $\{ \mathbf{C}^{i}_{seed} \}$ to obtain $\{ \mathbf{\hat{C}}^{i}_{seed} \}$. 
As the seed images $\{ \mathbf{C}^{i}_{seed} \}$ themselves may not be multiview-consistent, we first choose one view $\mathbf{\phi}^{0}_{seed}$ and inpaint the first seed image $\mathbf{\hat{C}}^{0}_{seed}$, and regard it as the seed of other additional seed images. This first inpainted image $\mathbf{\hat{C}}^{0}_{seed}$ is conditioned only on $\mathbf{C}^{0}_{seed}$ but nothing else. For other seed views $\mathbf{\phi}^{i}$ where $i > 0$, we use the stable diffusion model to inpaint $\mathbf{\hat{C}}^{i}_{seed}$, with the previously inpainted $\mathbf{\hat{C}}^{i-1}_{seed}$ as a strong prior. Details on how to utilize the previous inpainted image from a different camera as a prior will be introduced later in this subsection.

\vspace{0.5mm}
\noindent \textbf{Generating other views.}\quad
Once we have inpainted all the seed images $\{ \mathbf{\hat{C}}^{i}_{seed} \}$, we use them to interpolate all the other views to get the full set of pre-processed inpainted images $\{ \mathbf{\hat{C}}^{i} \}$. For a particular view $\mathbf{\phi}^{j}$, we choose its closest $n$ cameras (viewpoints) from $\{ \mathbf{\phi}^{i} \}$ to form $\{ \mathbf{\phi}^{1}, \mathbf{\phi}^{2}, \dots, \mathbf{\phi}^{n} \}$, associated with a set of weights $\{ w^1, w^2, \dots, w^n \}$ where the weights are proportional to the inverse of the camera distances and sum up to $1$:
\[
w^k = \frac{1}{(\mathbf{x}(\mathbf{\phi}^{k}) - \mathbf{x}(\mathbf{\phi}^{j})) \cdot \Sigma_{a=1}^{n} w^a}
\]
where $\mathbf{x}$ denotes the world space location of the viewpoints. These weights approximately reflect how much each neighboring view is related to the target view $\mathbf{\phi}^{j}$. According to these weights, we combine the visual content from the neighborhood cameras of $\mathbf{\phi}^{j}$ and generate an inpainted image for it. We use a projection-correction method to implement the aforementioned process as described below.

\vspace{0.5mm}
\noindent \textbf{Inpainted depth as geometry proxy.}\quad
To `copy' the inpainted content from $\mathbf{\phi}^{i}$ to $\mathbf{\phi}^{j}$, a simple approach is to cast rays from $\mathbf{\phi}^{i}$ into 3D space and then project them to $\mathbf{\phi}^{j}$, and the rgb values associated with the rays are rasterized onto $\mathbf{\phi}^{j}$. For this approach, in addition to the two camera poses, we need a depth map to determine which 3D locations the rays will cast to. The depth map is required to be plausible itself to make the projected content on $\mathbf{\phi}^{j}$ plausible. Nonetheless, it does not need to be perfectly accurate, since the purpose of the projection step is to give a rough prior on $\mathbf{\phi}^{j}$. Therefore, we assume a planar depth on $\mathbf{\phi}^{i}$. Specifically, on the depth map $\mathbf{D}^{i}$ of $\mathbf{\phi}^{i}$, we first perform pure inpainting using Lama~\cite{suvorov2021resolution} within the user-provided mask to erase the original depth, so that the depth within the mask is consistent with the background. Then, we feed the user's text prompt into LangSAM~\cite{liu2023grounding,kirillov2023segany} to generate an accurate mask of the inpainted object on $\mathbf{\hat{C}}^{i}$. Within this mask, the depth value is modified to be an appropriate constant for approximately accurate projections. Since we are substituting an object by another inpainted object, the original object's location is a good indicator of where the new object should locate. In this way, the inpainted object on $\mathbf{\hat{C}}^{i}$ is projected onto $\mathbf{\hat{C}}^{j}$ as if its geometry is planar, with the planar artifact to be auto-corrected later. In practice, to avoid projection artifacts on the background, we only keep the projection result within the mask of $\mathbf{\phi}^{j}$. Pixels outside the mask will remain the same.

For seed image generation from the first seed image, we only use 1 view for the projection. For generating non-seed views, we perform the projection $n$ times for the $n$ neighboring cameras respectively, which gives $n$ projected results. Multiplying these results with their associated weights $\{ w^1, w^2, \dots, w^n \}$, they are blended together as the final projection result. In practice, directly blending in image space may cause unnatural overlapping of different projected results. To reduce this artifact, we use the VAE of stable diffusion to encode the projections into latent tensors and perform blending in the latent space. Then the blended latent is decoded to be the final projection result. In this way, the stable diffusion correction step (introduced as follows) will have a higher chance to generate good results.

\vspace{0.5mm}
\noindent \textbf{Stable diffusion correction.}\quad
Since our planar depth estimation is not always accurate, while the raw projected results are plausible in general, the many small artifacts make the relevant images unqualified for training. To make them look more natural, we propose to cover the projection artifacts with stable diffusion hallucinated details. First, we blend the raw projection with random noise in stable diffusion's latent space for $t$ timesteps, where $t$ is relatively small to the total number of stable diffusion timesteps. Then, starting from the last $t$ steps, we denoise with stable diffusion to generate the hallucinated image. The resulting image is then regarded as the initial training image. With this step, the training image pre-processing stage is completed.


\subsection{Progressive Training}

\vspace{0.5mm}
\noindent \textbf{Warmup Training.}\quad
Our training image pre-processing stage provides a good initialization for rough convergence. Before fine-tuning on these images to get fine convergence, notice that the 3D object in NeRF is still the original object intact, which is quite different from what is depicted in the pre-processed training images above in geometry and appearance. Therefore, for the first stage of NeRF training, without any finetuning, we directly train the NeRF on the pre-processed images for coarse convergence. Note that when we perform 
fine-tuning later, the original object will have no effect because warmup training has already erased all its appearance information.

\vspace{0.5mm}
\noindent \textbf{Iterative Dataset Update.}\quad
Given a NeRF in which the editing target has fixed overall geometry, Iterative Dataset Update (IDU), first proposed and shown effective in Instruct-Nerf2Nerf~\cite{haque2023instruct}, is a useful training image fine-tuning strategy which can edit the appearance and fine geometry of the target object in NeRF. In our task, warmup training has already provided a converged coarse geometry, leaving fine geometry and appearance to be determined. In their work, they use InstructPix2Pix~\cite{brooks2022instructpix2pix} as a diffusion model strongly conditioned on the original training image. In our task, we use stable diffusion in a similar way to achieve similar objectives. The fine-tuned image should be conditioned on the pre-processed image, as well as the depth map. In detail, for pre-processed image conditioning, we adopt the same approach as in the previously introduced projected image correction. The preprocessed image and the current NeRF rendering are blended together, either in image space or in latent space. Then a small amount of noise is injected into the blended image, followed by stable diffusion denoising from an intermediate timestep. For depth map conditioning, we input the depth map of the current NeRF into ControlNet to guide the editing process.

Along with training image updates and NeRF training, the underlying NeRF will gradually converge to an object with fine details, making rendering from NeRF more reliable. Meanwhile, stable diffusion correction with less noise injected into the training images introduces less multiview inconsistency to fine details. Therefore, we gradually increase the intermediate timestep of our stable diffusion correction when training NeRF, until the intermediate timestep becomes very close to the final denoise timestep. Consequently, the coarse NeRF converges to a fine one.

\vspace{0.5mm}
\noindent \textbf{Regularizers.}\quad
To supervise NeRF training, we use the L1 photometric loss between NeRF rendering and the training images. However, sole supervision of RGB cannot avoid inaccuracies and noise in geometry. 
To achieve clean converged geometry, we add the following two regularizers into supervision. First, 
we also render a depth value along a ray, and compute the depth loss as defined in~\cite{kangle2021dsnerf} between the rendered depth and our pre-processed 2-layer depth i.e., initial inpainted geometry proxies. Supervision with our planar 2-layer depth is not accurate, but it helps to avoid incorrect converged geometry, such as merging of the inpainted geometry with the background. Second, to deal with noisy RGB and depth renderings and ``floaters" (artifacts hovering in the underlying NeRF volume), we use LPIPS loss~\cite{zhang2018perceptual} on RGB as a regularizing term. 
In practice, we found LPIPS to be effective in reducing noise and floaters. 
Since LPIPS is based on patches, rays selected in NeRF training are not allowed to be random, but are required to form rectangular patches. Thus we render small image patches, corresponding to rectangular areas in ground truth images, during NeRF optimization. The effect of the regularizers is also previously demonstrated in ~\cite{Xian2020SpacetimeNI, inpaintnerf360}.

\subsection{4D Extension}
Our method extends naturally to 4D by adopting the idea in 3D, that is, to have a raw inference from the seed image, and correct its details with stable diffusion. To infer across frames, we apply multiple existing video editing methods to the input video. We use point-tracking to record the original movement. This movement will be applied to animate the novel object generated through our progressive training. Meanwhile, the foreground of the original video will be removed to adjust the new scene. The video with only the background will be combined with the animated novel object to build up a raw inpainted video that is temporally consistent. For each frame, we extract the seed image from this seed video and project it to other views. Finally, training images from all views and frames are refined by stable diffusion before being used in training.

To determine the motion for the new object, since there exists no stable method for generating a background-consistent movement for a generated foreground object, we utilize the original movement of the replaced object. To achieve this goal, it is required to first extract the original movement, and then apply it back to the novel object.
We use CoTracker~\cite{karaev2023cotracker} to track the movements of multiple key points on the target. CoTracker enables us to track self-defined key points on the original video precisely. Making use of this functionality, the trajectory followed by the key points can be obtained and further projected to the generated object with consistent background information. In this way, we are able to animate the generated object.

\section{Experiments}
\begin{figure*}[htbp]
    \begin{minipage}[b]{0.2\textwidth}
        \centering
        Initial rgb\\~\\~
    \end{minipage}
    \hfill
    \begin{subfigure}[b]{0.25\textwidth}
         \centering
         \includegraphics[width=\textwidth]{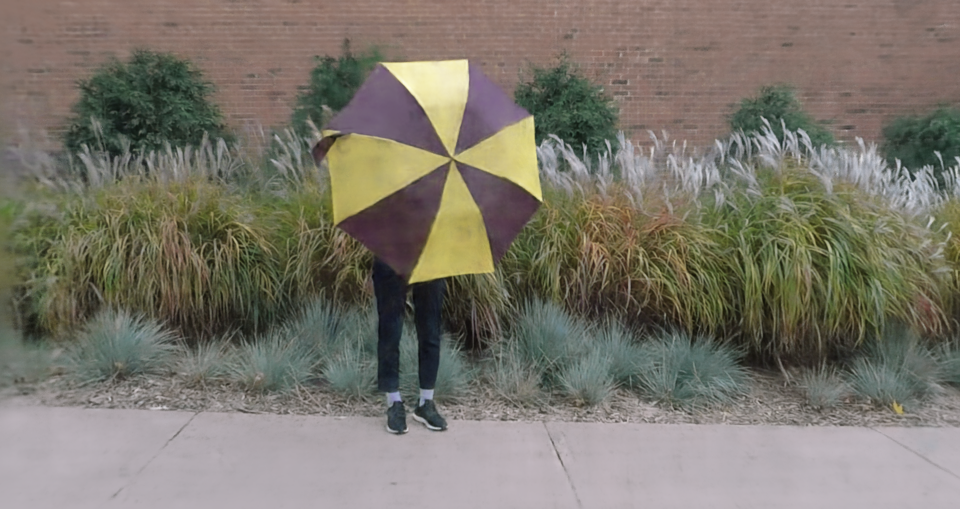}
    \end{subfigure}
    \hfill
    \begin{subfigure}[b]{0.25\textwidth}
         \centering
         \includegraphics[width=\textwidth]{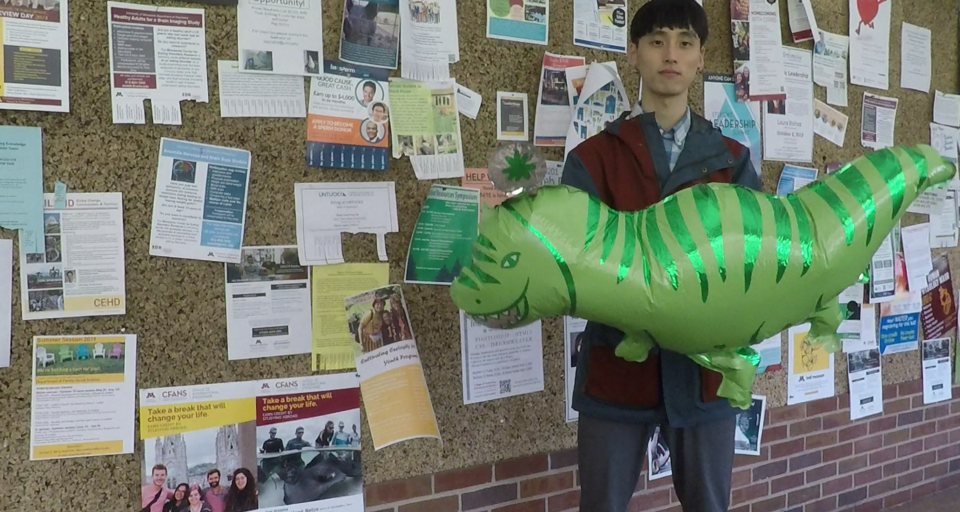}
    \end{subfigure}
    \hfill
    \begin{subfigure}[b]{0.25\textwidth}
         \centering
         \includegraphics[width=\textwidth]{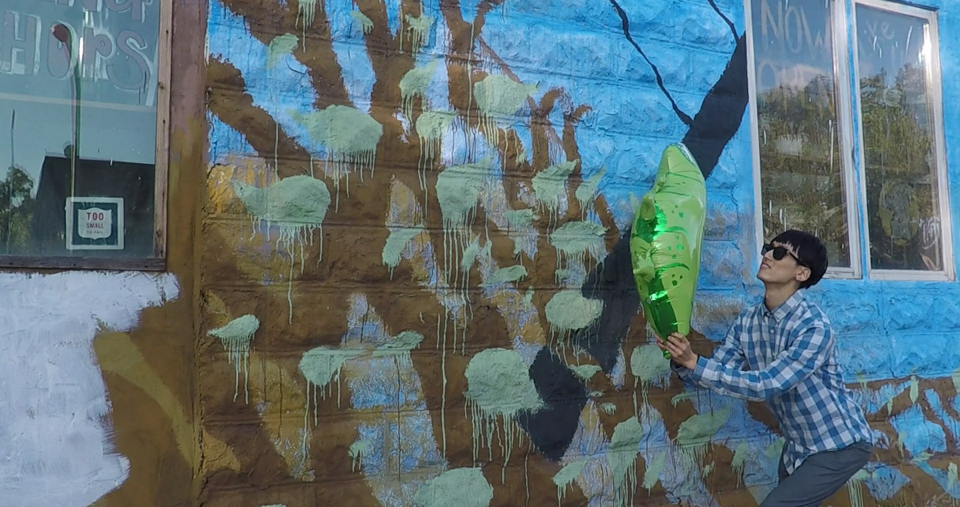}
    \end{subfigure}
    \hfill
    \begin{minipage}[b]{0.2\textwidth}
        \centering
        Initial depth\\~\\~
    \end{minipage}
    \hfill
    \begin{subfigure}[b]{0.25\textwidth}
         \centering
         \includegraphics[width=\textwidth]{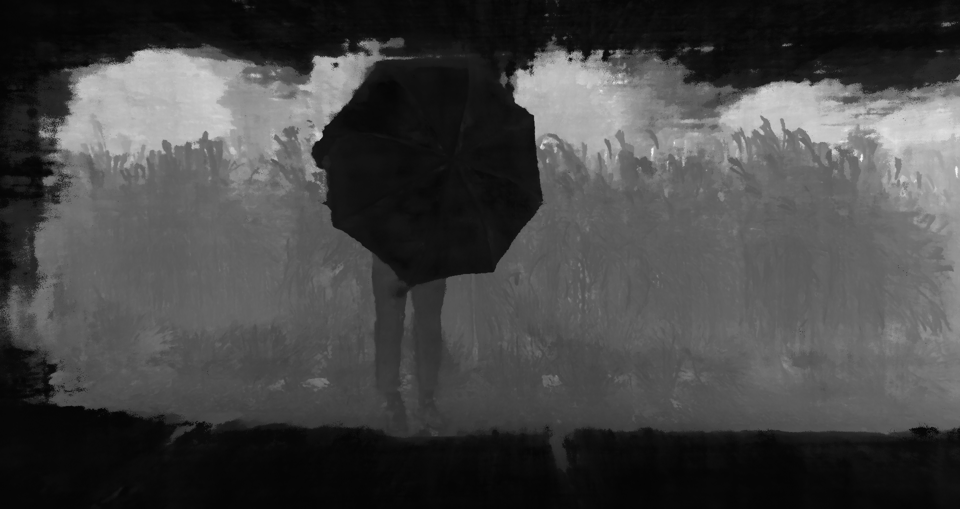}
    \end{subfigure}
    \hfill
    \begin{subfigure}[b]{0.25\textwidth}
         \centering
         \includegraphics[width=\textwidth]{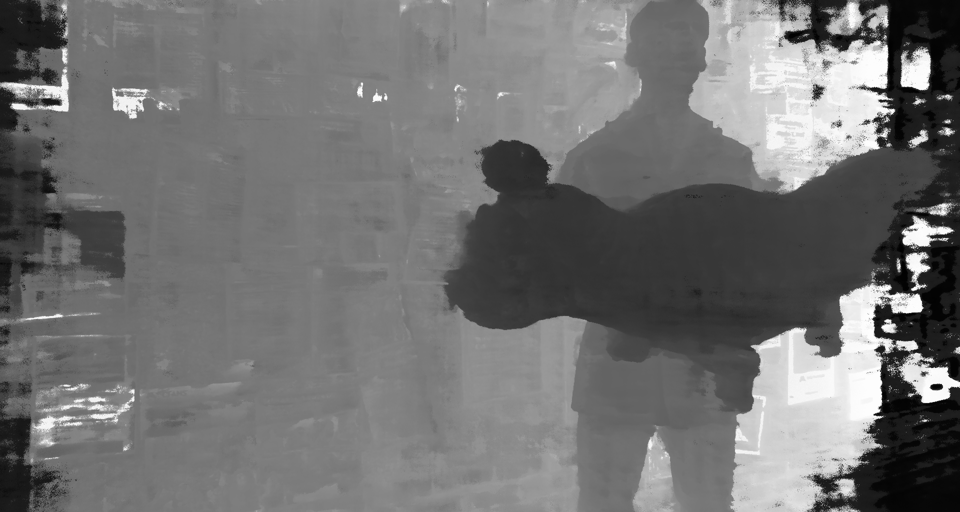}
    \end{subfigure}
    \hfill
    \begin{subfigure}[b]{0.25\textwidth}
         \centering
         \includegraphics[width=\textwidth]{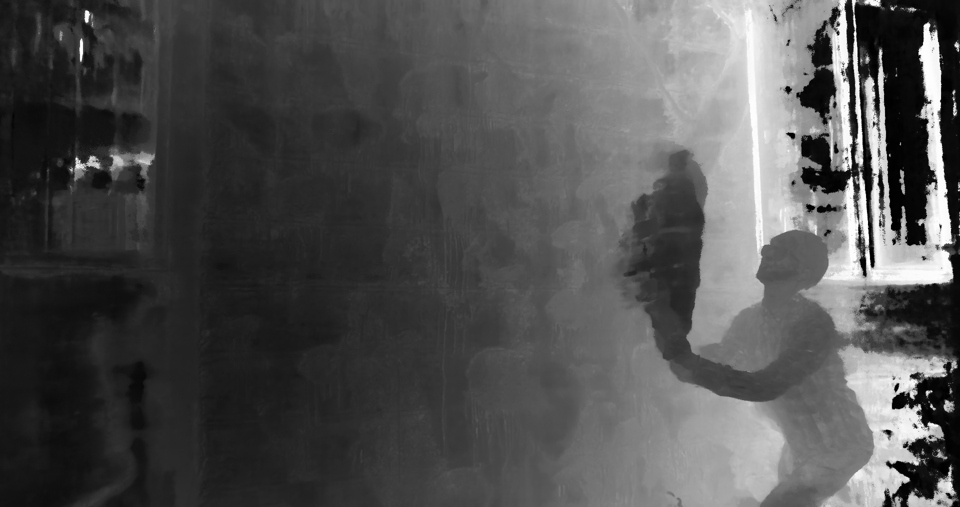}
    \end{subfigure}
    \hfill
    \begin{minipage}[b]{0.2\textwidth}
        \centering
        First seed \\~\\~
    \end{minipage}
    \hfill
    \begin{subfigure}[b]{0.25\textwidth}
         \centering
         \includegraphics[width=\textwidth]{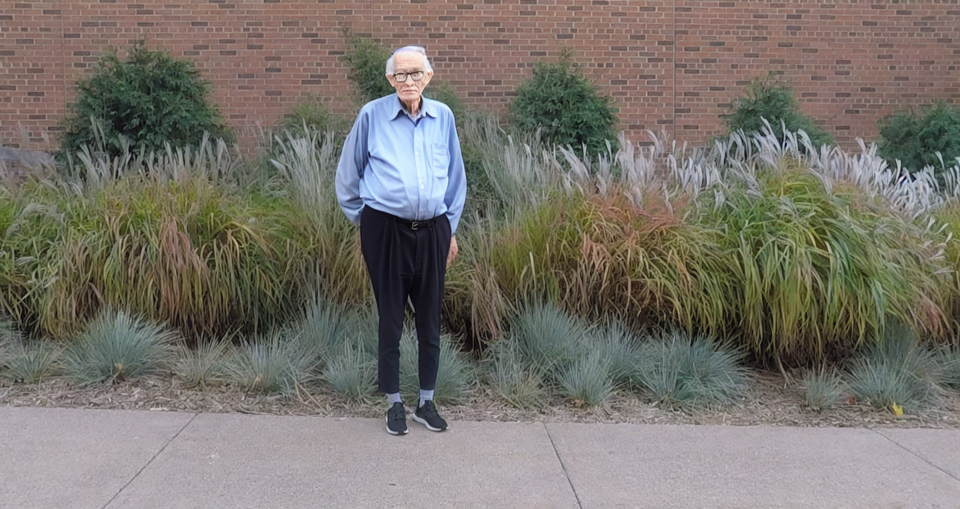}
    \end{subfigure}
    \hfill
    \begin{subfigure}[b]{0.25\textwidth}
         \centering
         \includegraphics[width=\textwidth]{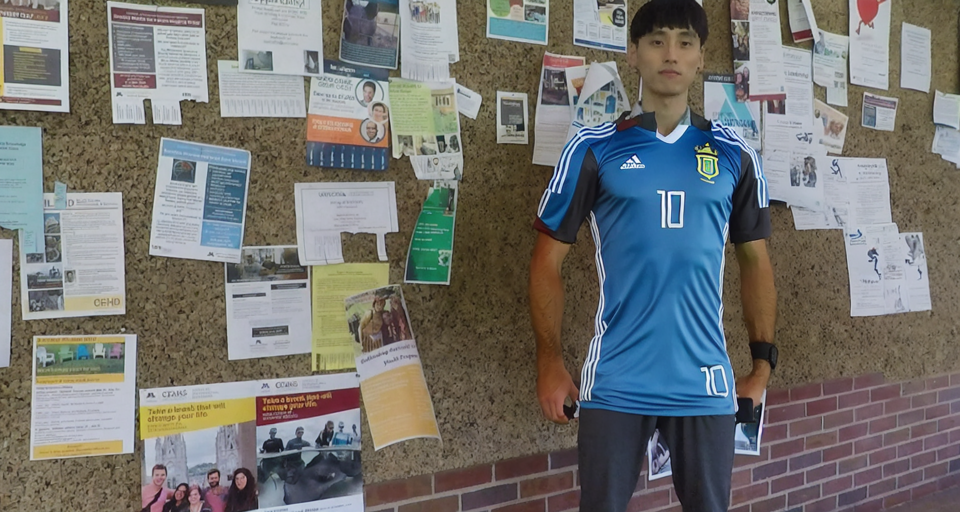}
    \end{subfigure}
    \hfill
    \begin{subfigure}[b]{0.25\textwidth}
         \centering
         \includegraphics[width=\textwidth]{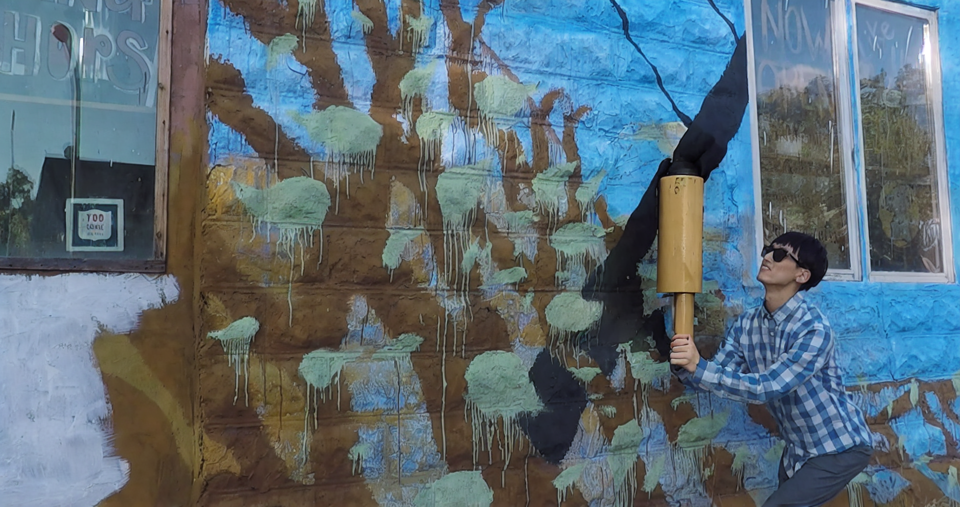}
    \end{subfigure}
    \hfill
    \begin{minipage}[b]{0.2\textwidth}
        \centering
        Another preprocessed image \\~\\~
    \end{minipage}
    \hfill
    \begin{subfigure}[b]{0.25\textwidth}
         \centering
         \includegraphics[width=\textwidth]{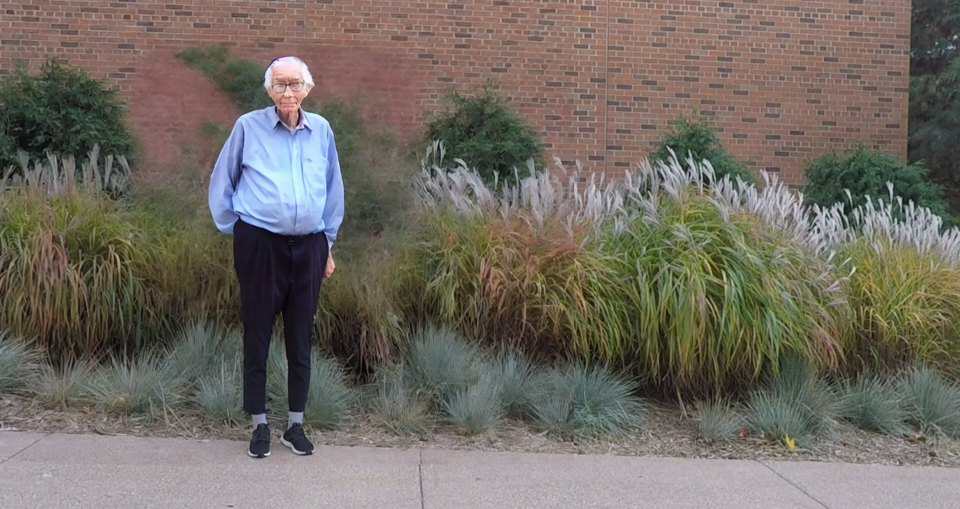}
    \end{subfigure}
    \hfill
    \begin{subfigure}[b]{0.25\textwidth}
         \centering
         \includegraphics[width=\textwidth]{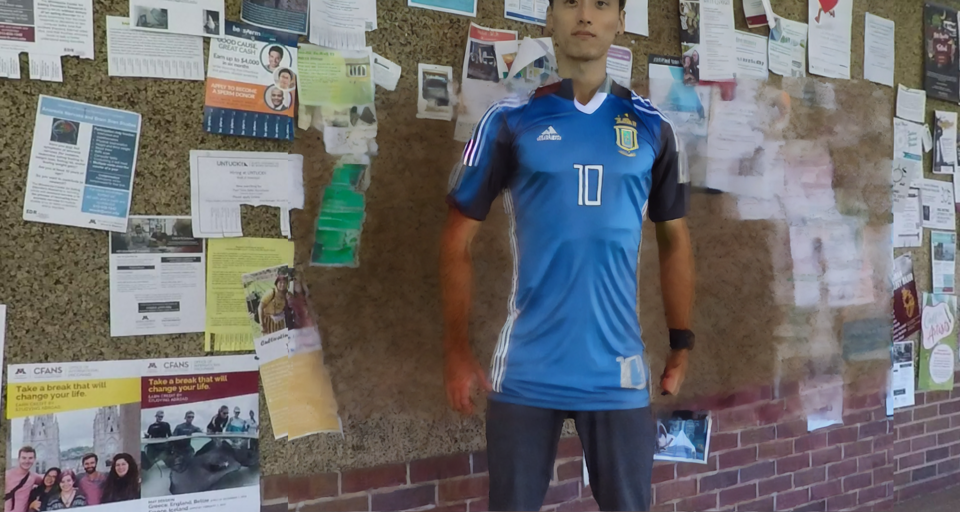}
    \end{subfigure}
    \hfill
    \begin{subfigure}[b]{0.25\textwidth}
         \centering
         \includegraphics[width=\textwidth]{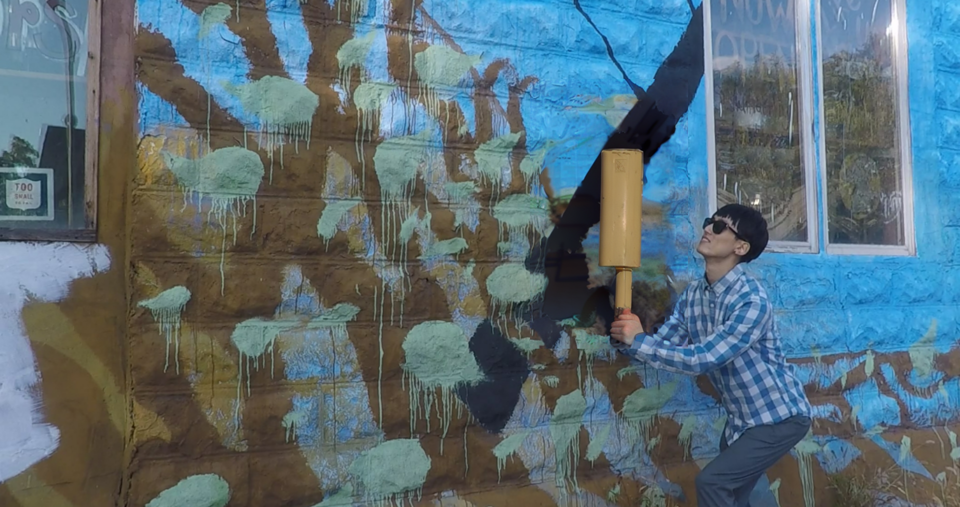}
    \end{subfigure}
    \hfill
    \begin{minipage}[b]{0.2\textwidth}
        \centering
        Warmup rgb\\~\\~
    \end{minipage}
    \hfill
    \begin{subfigure}[b]{0.25\textwidth}
         \centering
         \includegraphics[width=\textwidth]{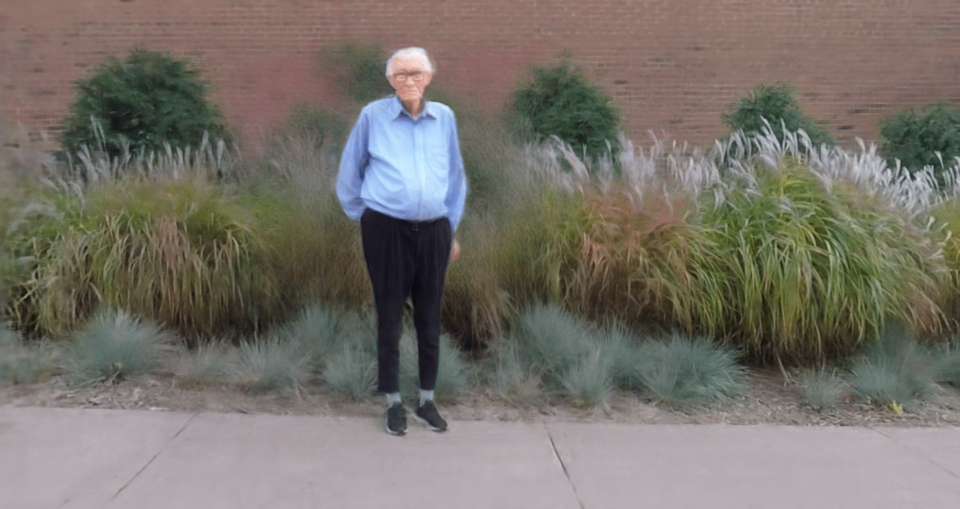}
    \end{subfigure}
    \hfill
    \begin{subfigure}[b]{0.25\textwidth}
         \centering
         \includegraphics[width=\textwidth]{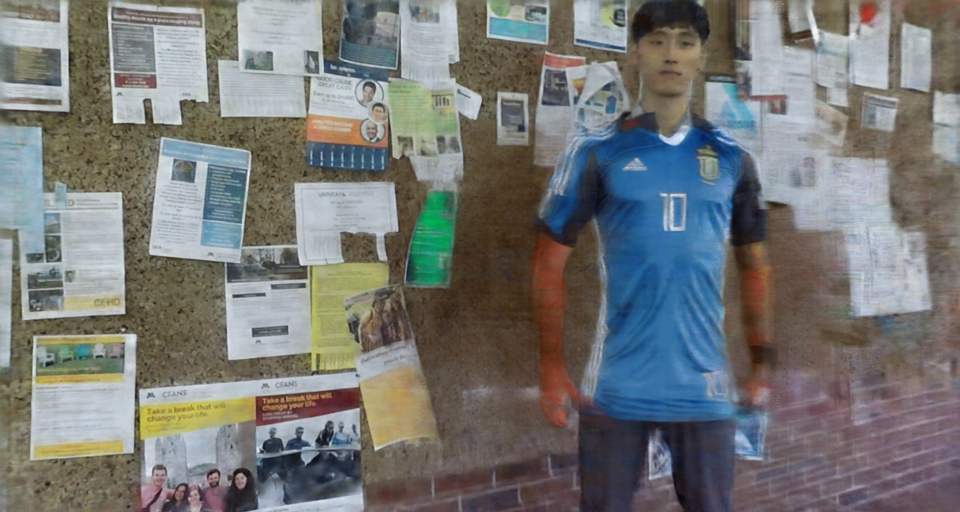}
    \end{subfigure}
    \hfill
    \begin{subfigure}[b]{0.25\textwidth}
         \centering
         \includegraphics[width=\textwidth]{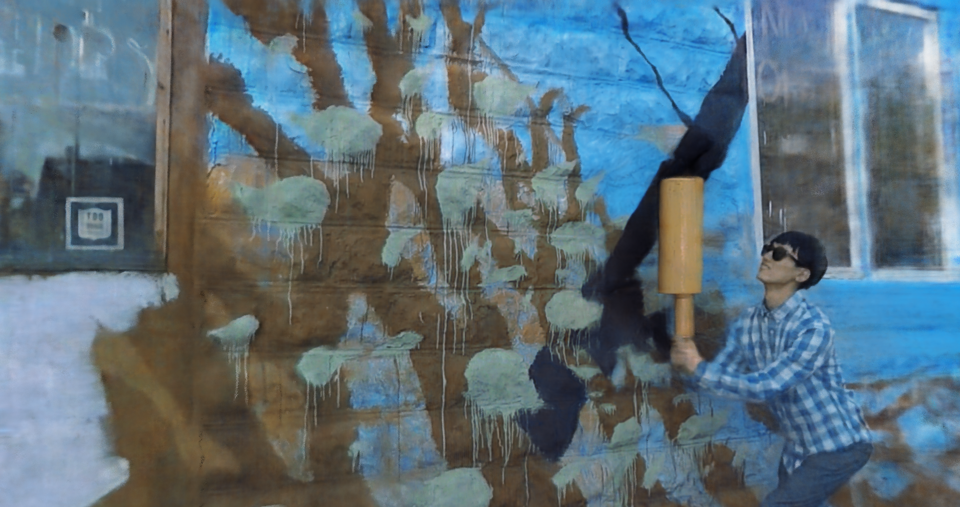}
    \end{subfigure}
    \hfill
    \begin{minipage}[b]{0.2\textwidth}
        \centering
        Warmup Depth\\~\\~
    \end{minipage}
    \hfill
    \begin{subfigure}[b]{0.25\textwidth}
         \centering
         \includegraphics[width=\textwidth]{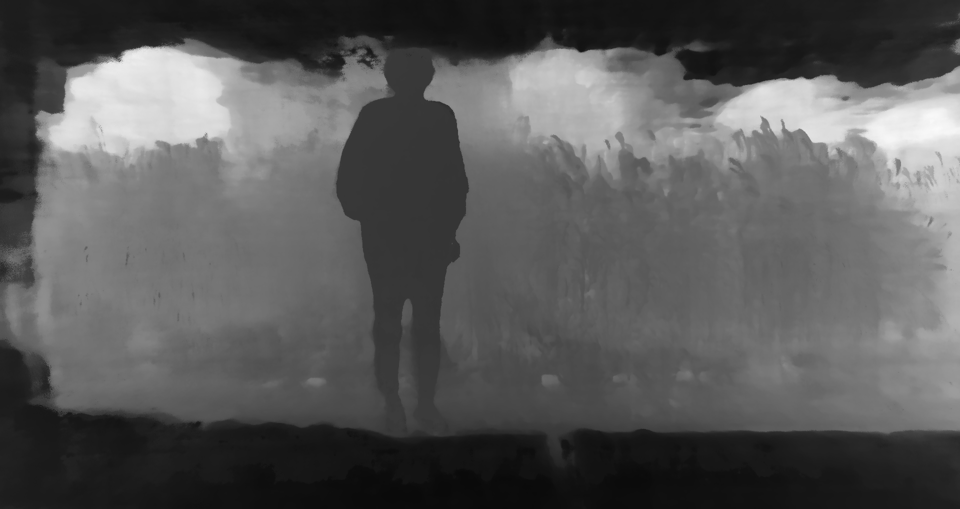}
    \end{subfigure}
    \hfill
    \begin{subfigure}[b]{0.25\textwidth}
         \centering
         \includegraphics[width=\textwidth]{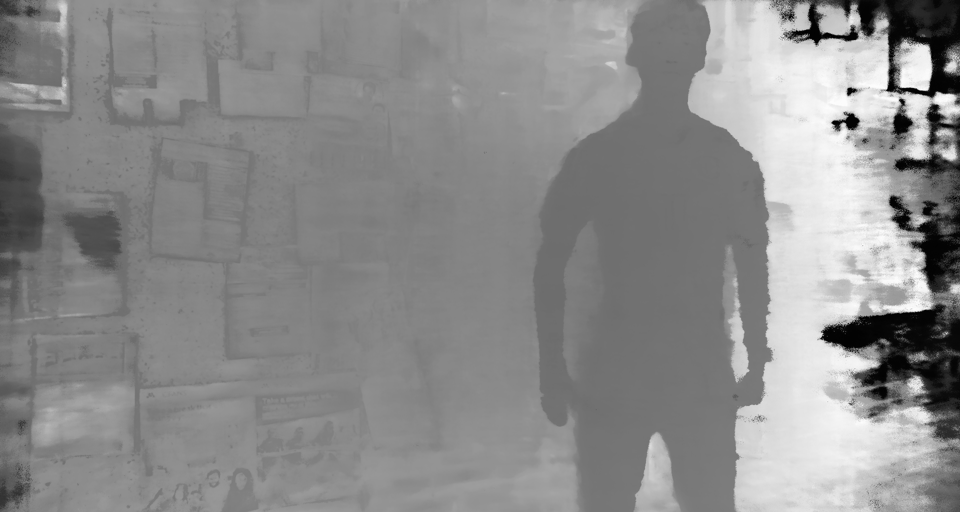}
    \end{subfigure}
    \hfill
    \begin{subfigure}[b]{0.25\textwidth}
         \centering
         \includegraphics[width=\textwidth]{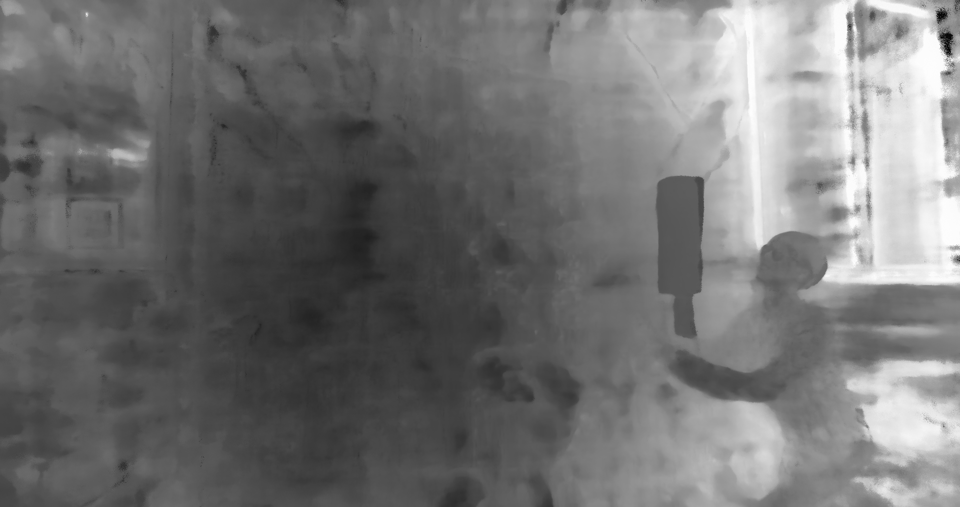}
    \end{subfigure}
    \hfill
    \begin{minipage}[b]{0.2\textwidth}
        \centering
        Final rgb\\~\\~
    \end{minipage}
    \hfill
    \begin{subfigure}[b]{0.25\textwidth}
         \centering
         \includegraphics[width=\textwidth]{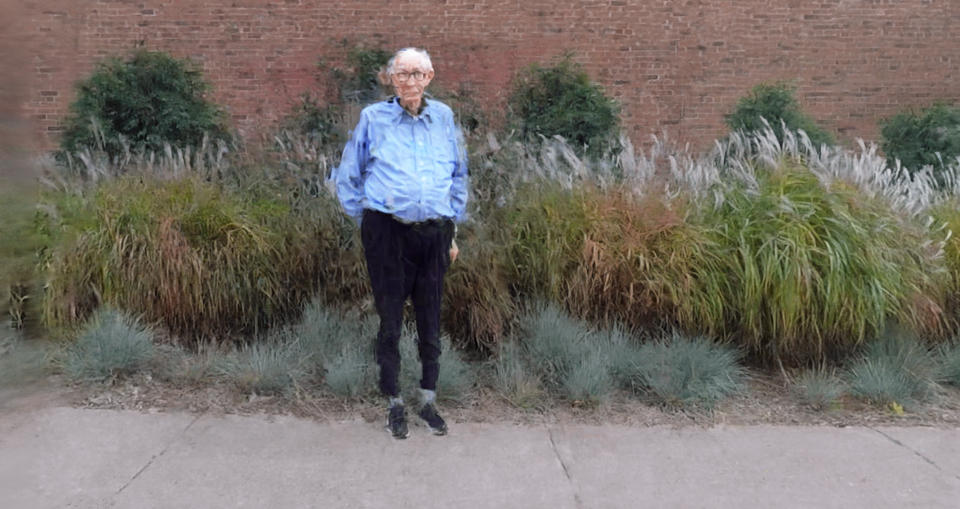}
    \end{subfigure}
    \hfill
    \begin{subfigure}[b]{0.25\textwidth}
         \centering
         \includegraphics[width=\textwidth]{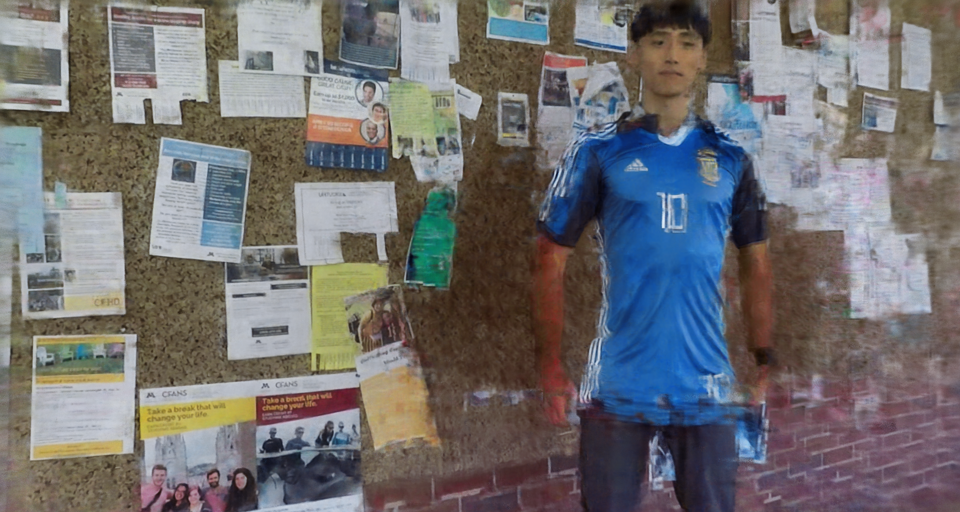}
    \end{subfigure}
    \hfill
    \begin{subfigure}[b]{0.25\textwidth}
         \centering
         \includegraphics[width=\textwidth]{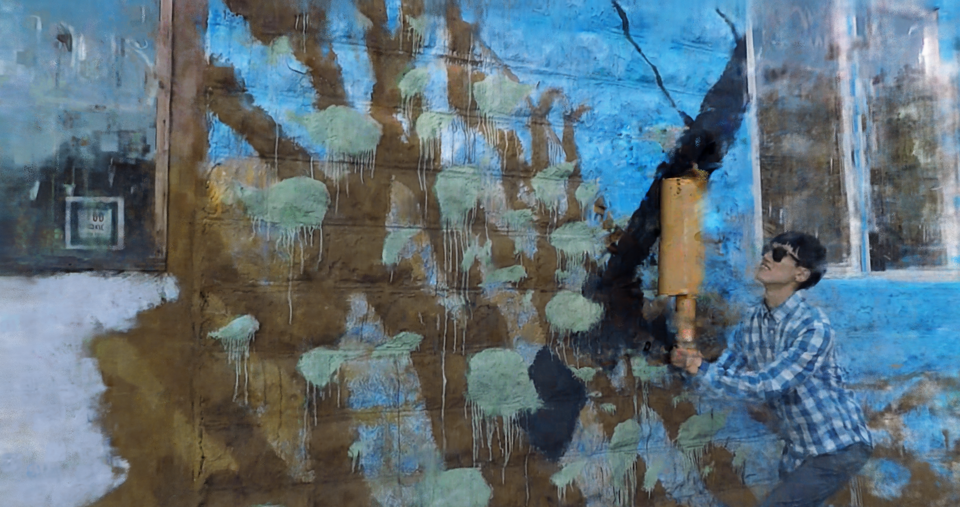}
    \end{subfigure}
    \hfill
    \begin{minipage}[b]{0.2\textwidth}
        \centering
        Final depth\\~\\~
    \end{minipage}
    \hfill
    \begin{subfigure}[b]{0.25\textwidth}
         \centering
         \includegraphics[width=\textwidth]{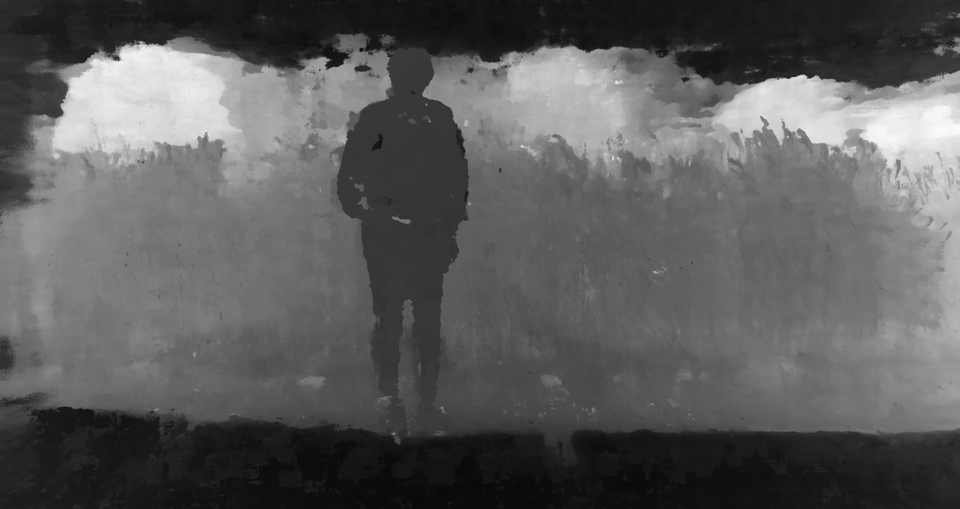}
    \end{subfigure}
    \hfill
    \begin{subfigure}[b]{0.25\textwidth}
         \centering
         \includegraphics[width=\textwidth]{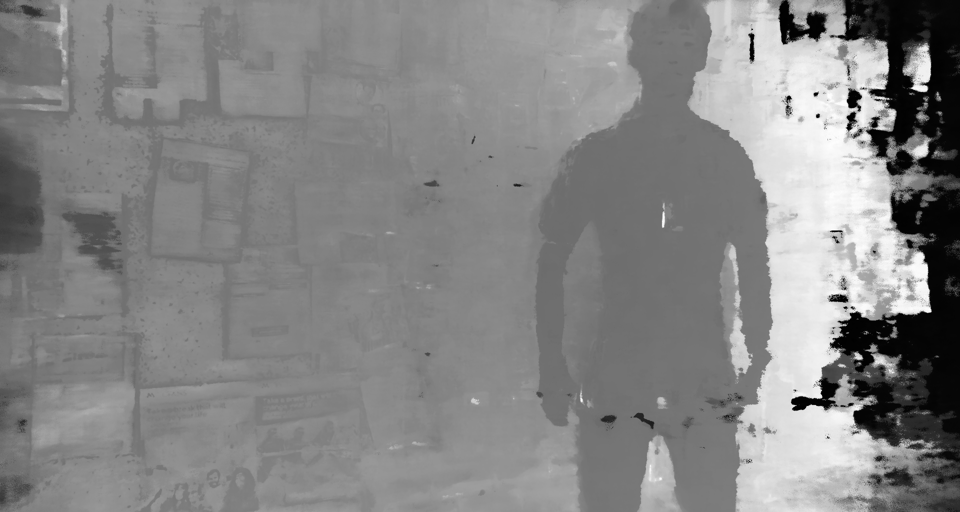}
    \end{subfigure}
    \hfill
    \begin{subfigure}[b]{0.25\textwidth}
         \centering
         \includegraphics[width=\textwidth]{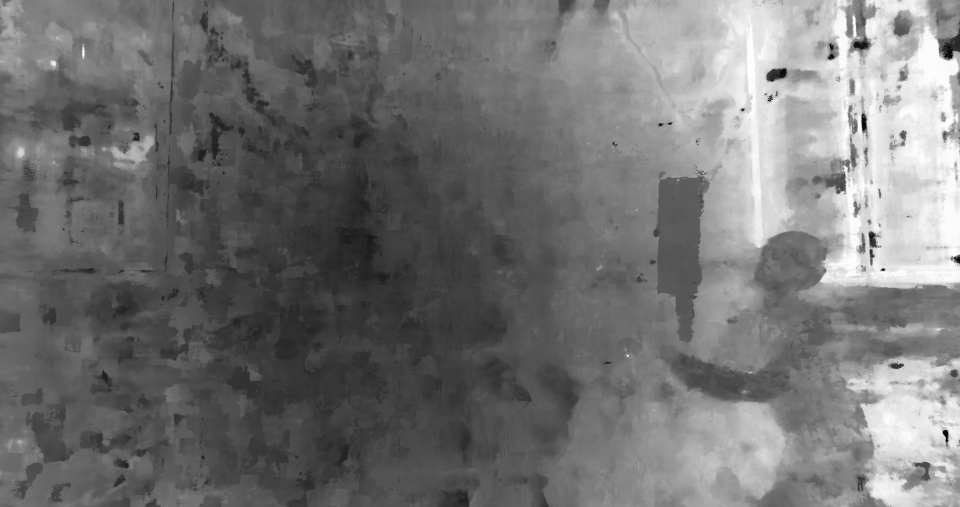}
    \end{subfigure}
    \hfill
    \begin{minipage}[b]{0.2\textwidth}
        \centering
        Final rgb, another view\\~\\~\\~
    \end{minipage}
    \hfill
    \begin{subfigure}[b]{0.25\textwidth}
         \centering
         \includegraphics[width=\textwidth]{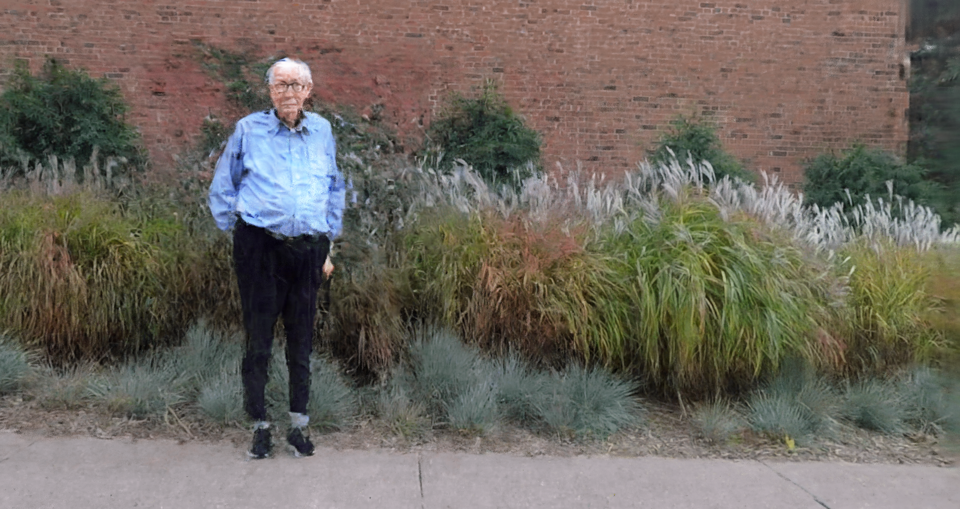}
         \caption{``an old professor, standing''}
    \end{subfigure}
    \hfill
    \begin{subfigure}[b]{0.25\textwidth}
         \centering
         \includegraphics[width=\textwidth]{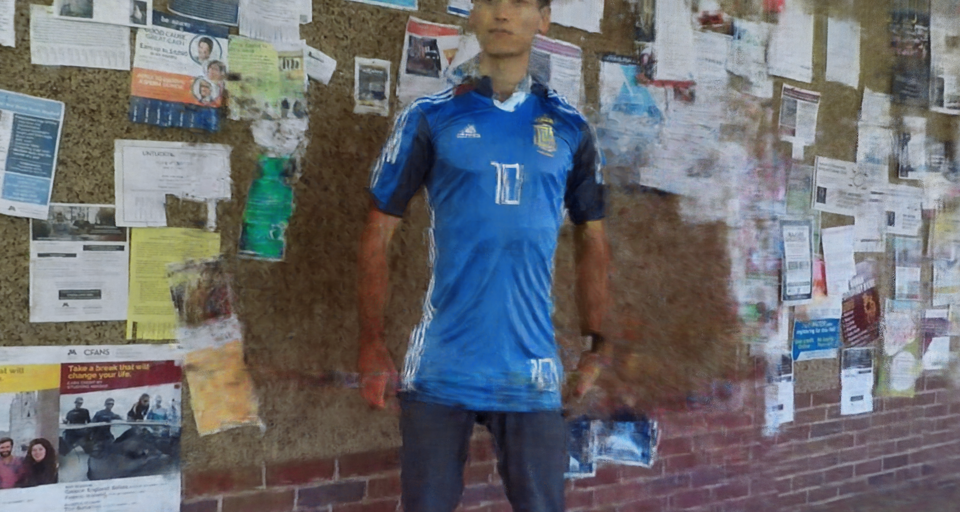}
         \caption{``an Argentina soccer player''}
    \end{subfigure}
    \hfill
    \begin{subfigure}[b]{0.25\textwidth}
         \centering
         \includegraphics[width=\textwidth]{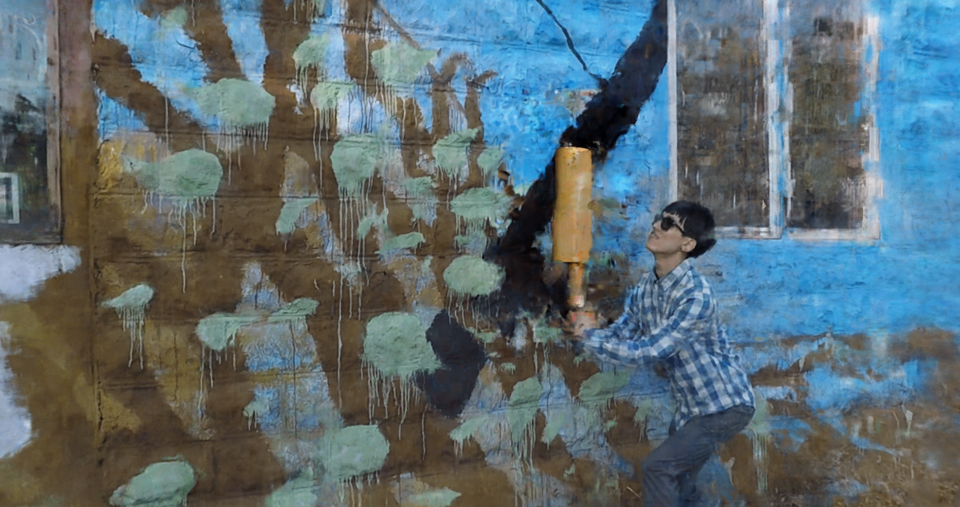}
         \caption{``telegraph pole''}
    \end{subfigure}
\caption{Our qualitative results in 3D. Each column illustrates an inpainting example. We show final renderings from 2 views to demonstrate the multiview consistency. We also show depth maps and rgb images of different training stages to show their roles.}
\label{fig:qual}
\end{figure*}
\label{sec:experiments}

For our experiments, we select from dynamic scenes in the Nvidia Dynamic Scenes Dataset~\cite{nvidia}. Scenes in this dataset are captured using a sparse set of 12 stationary cameras located in two rows, producing images of resolution 1015$\times$1920. The static scenes we use are taken from one frame from the dynamic scenes. For the backbone NeRF, we use static and dynamic versions of K-Planes~\cite{kplanes_2023} implemented in nerfstudio~\cite{nerfstudio}. For each scene, we conduct inpainting by replacing a foreground object with another text-prompted object with a different geometry. We will demonstrate the effectiveness of our method by showing the qualitative intermediate and final results. In addition, we will explain different parts of our design by ablations and comparisons on our baseline.

\subsection{Qualitative results}

\vspace{0.5mm}
\noindent \textbf{3D Examples.}\quad
We show several 3D inpainting examples in figure~\ref{fig:qual}. For each individual inpainting task, we show 2 renderings of the final NeRF from different views to demonstrate the multiview consistency. Additionally, we show the first seed image, another pre-processed image, as well as the RGB and depth map in the three stages: before training, after warmup training, and after convergence. These before-and-after images demonstrate the efficacy of each stage in our method. As shown in Figure~\ref{fig:qual}, a roughly consistent pre-processed image can optimize a coarse inpainted NeRF after warmup training, and the geometry (represented by depth map) converges during warmup training. Then, fine convergence across views is achieved after the final training stage. All 3D inpainting tasks are trained on a single Nvidia RTX 4090 GPU. Warmup training takes approximately 0.5--1 hour, and the main training stage with IDU takes approximately 1--2 hours.

\begin{figure*}[htbp]
    \begin{minipage}[b]{0.2\textwidth}
        \centering
        Original \\~\\~
    \end{minipage}
    \hfill
    \begin{subfigure}[b]{0.25\textwidth}
         \centering
         \includegraphics[width=\textwidth]{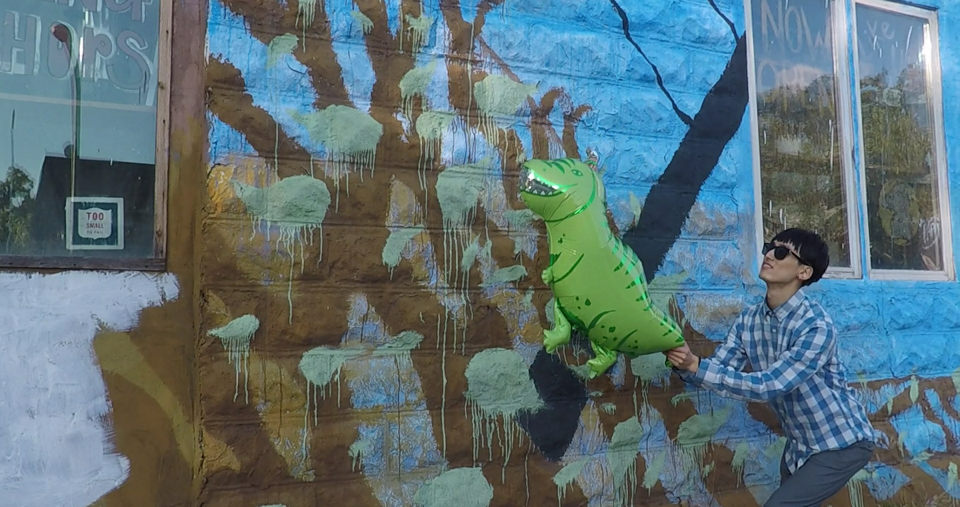}
    \end{subfigure}
    \hfill
    \begin{subfigure}[b]{0.25\textwidth}
         \centering
         \includegraphics[width=\textwidth]{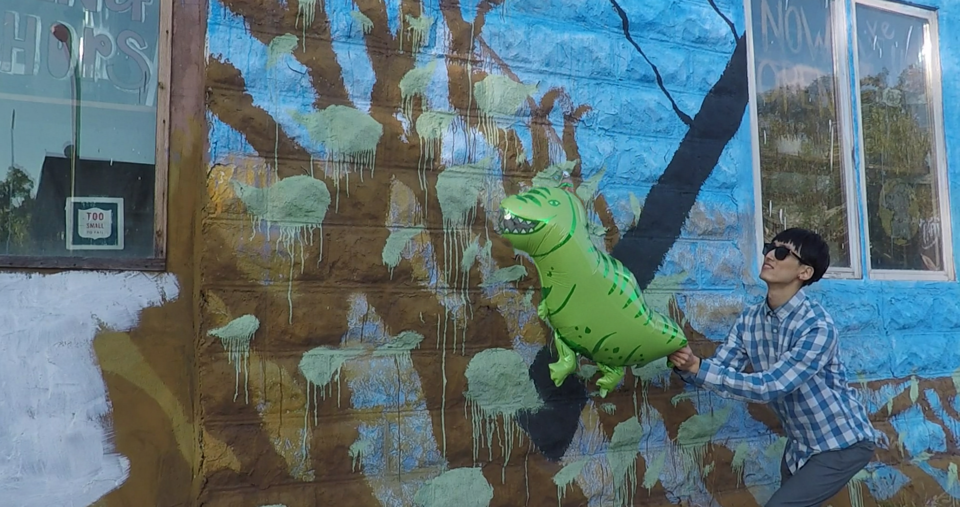}
    \end{subfigure}
    \hfill
    \begin{subfigure}[b]{0.25\textwidth}
         \centering
         \includegraphics[width=\textwidth]{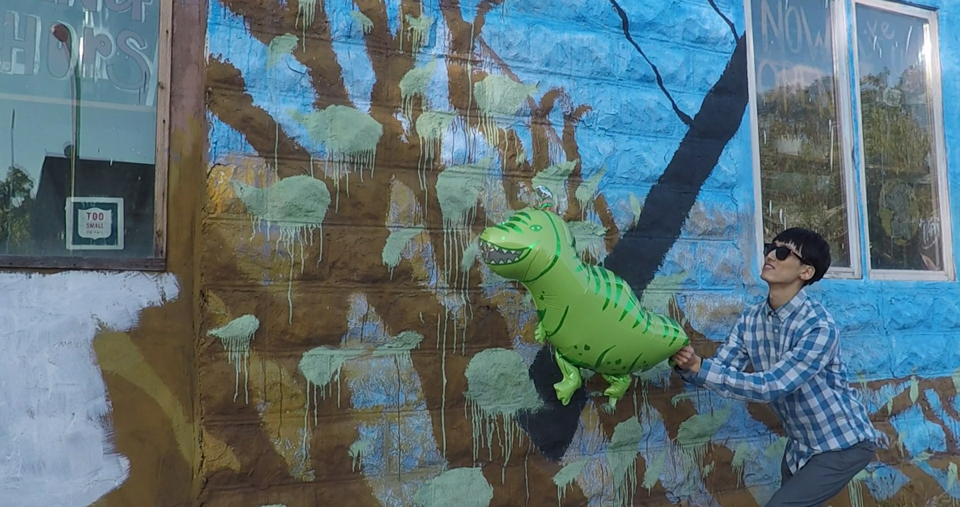}
    \end{subfigure}
    \begin{minipage}[b]{0.2\textwidth}
        \centering
        Seed video \\~\\~
    \end{minipage}
    \hfill
    \begin{subfigure}[b]{0.25\textwidth}
         \centering
         \includegraphics[width=\textwidth]{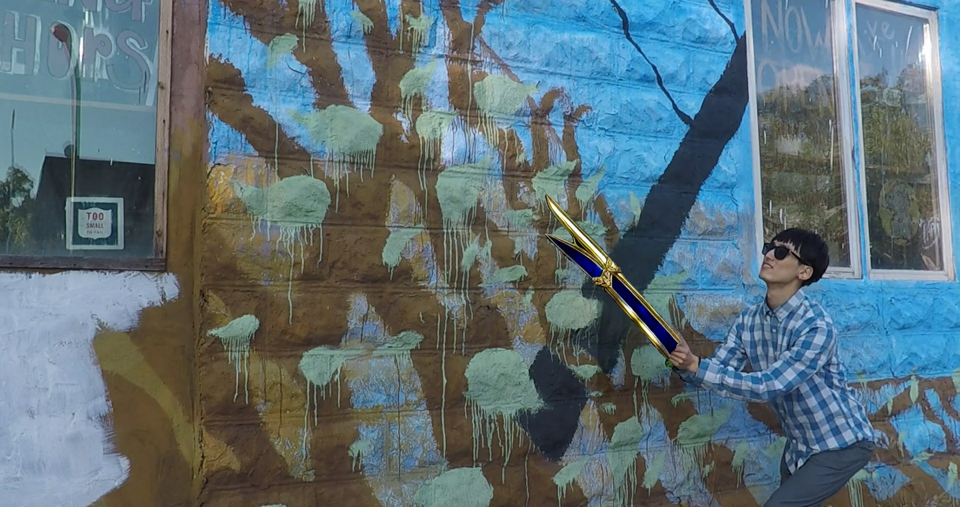}
    \end{subfigure}
    \hfill
    \begin{subfigure}[b]{0.25\textwidth}
         \centering
         \includegraphics[width=\textwidth]{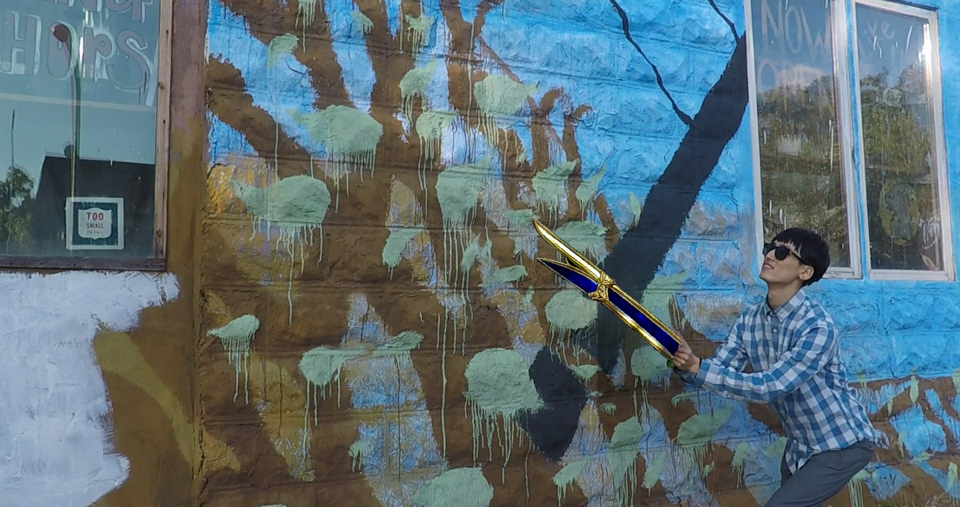}
    \end{subfigure}
    \hfill
    \begin{subfigure}[b]{0.25\textwidth}
         \centering
         \includegraphics[width=\textwidth]{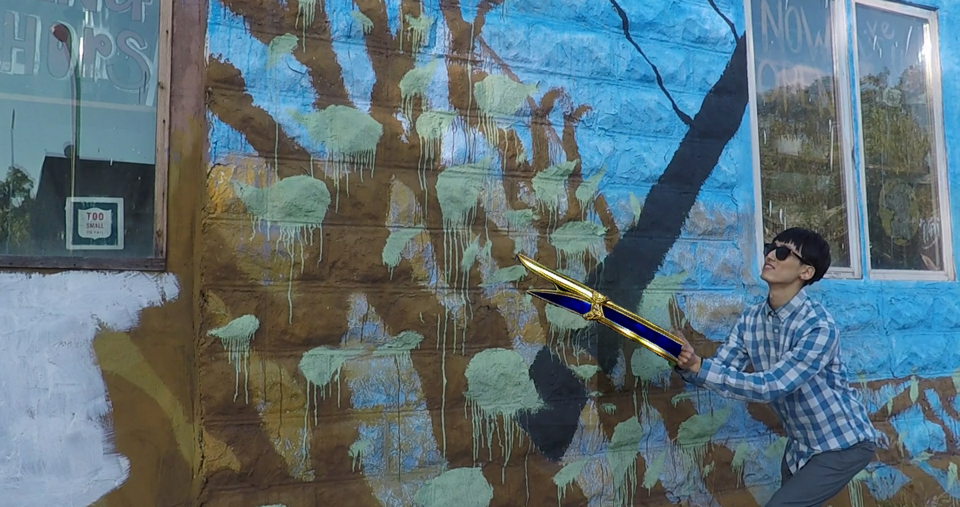}
    \end{subfigure}
    \begin{minipage}[b]{0.2\textwidth}
        \centering
        Final NeRF render \\~\\~
    \end{minipage}
    \hfill
    \begin{subfigure}[b]{0.25\textwidth}
         \centering
         \includegraphics[width=\textwidth]{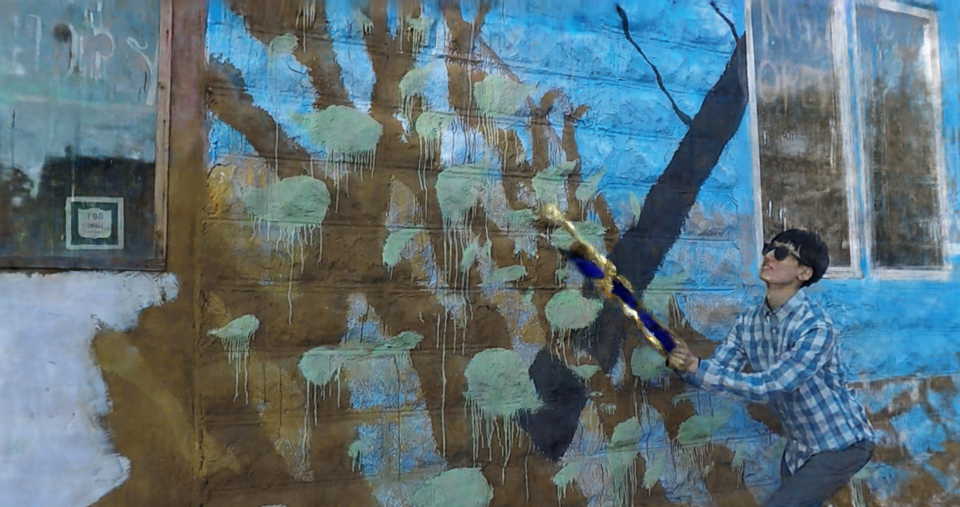}
    \end{subfigure}
    \hfill
    \begin{subfigure}[b]{0.25\textwidth}
         \centering
         \includegraphics[width=\textwidth]{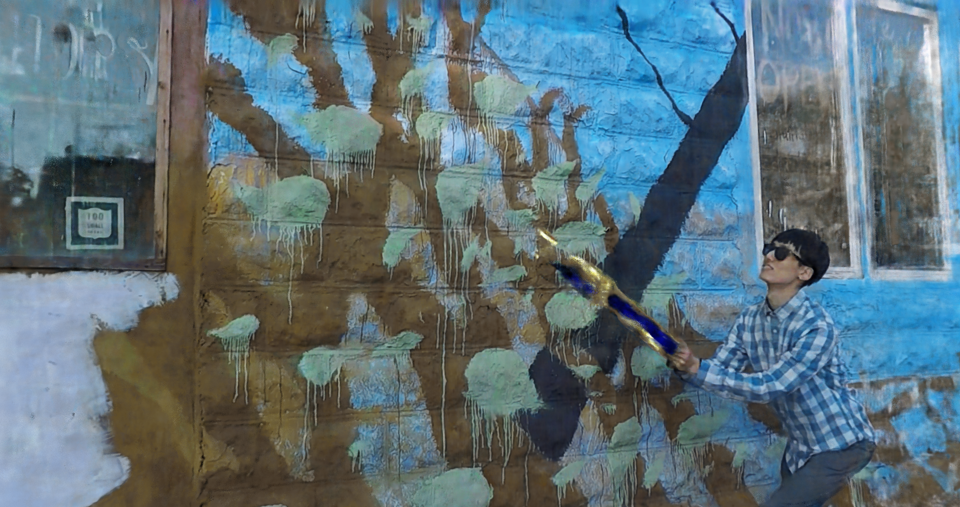}
    \end{subfigure}
    \hfill
    \begin{subfigure}[b]{0.25\textwidth}
         \centering
         \includegraphics[width=\textwidth]{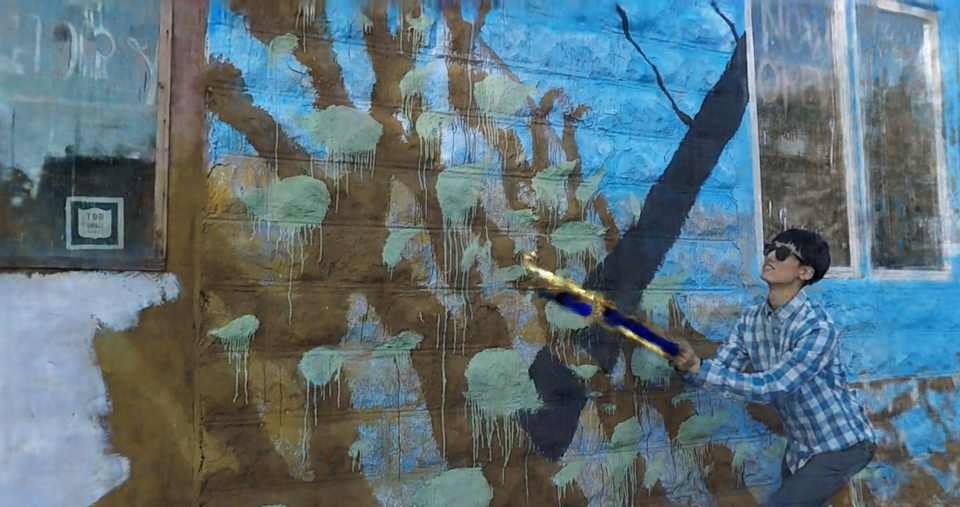}
    \end{subfigure}
    \begin{minipage}[b]{0.2\textwidth}
        \centering
        Final NeRF depth map \\~\\~
    \end{minipage}
    \hfill
    \begin{subfigure}[b]{0.25\textwidth}
         \centering
         \includegraphics[width=\textwidth]{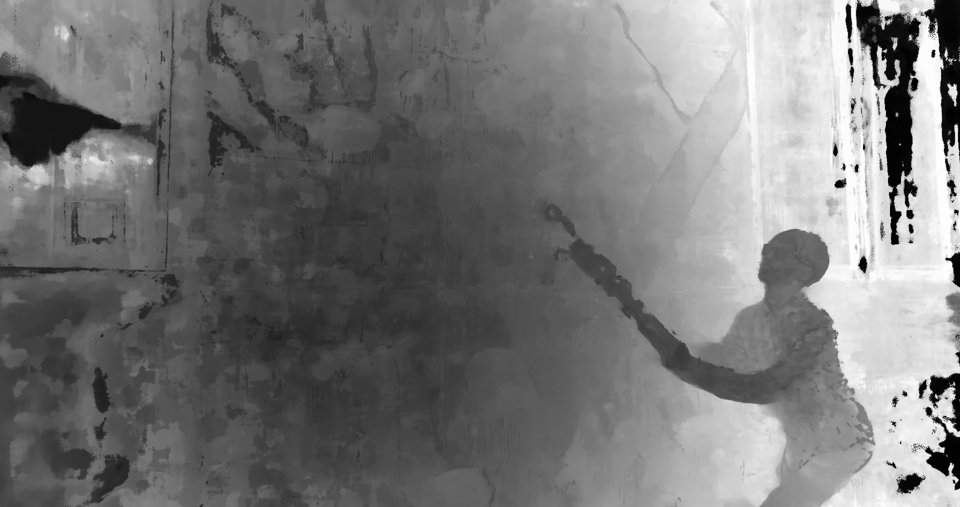}
    \end{subfigure}
    \hfill
    \begin{subfigure}[b]{0.25\textwidth}
         \centering
         \includegraphics[width=\textwidth]{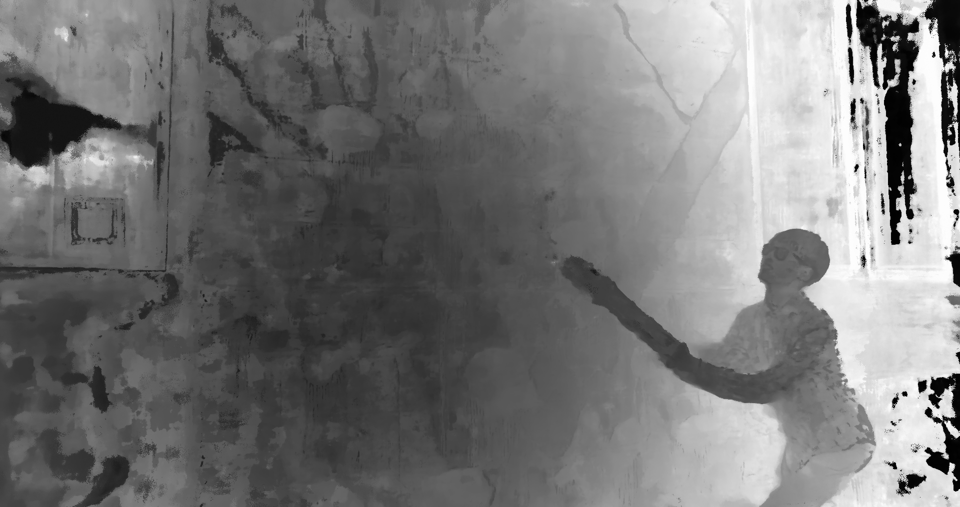}
    \end{subfigure}
    \hfill
    \begin{subfigure}[b]{0.25\textwidth}
         \centering
         \includegraphics[width=\textwidth]{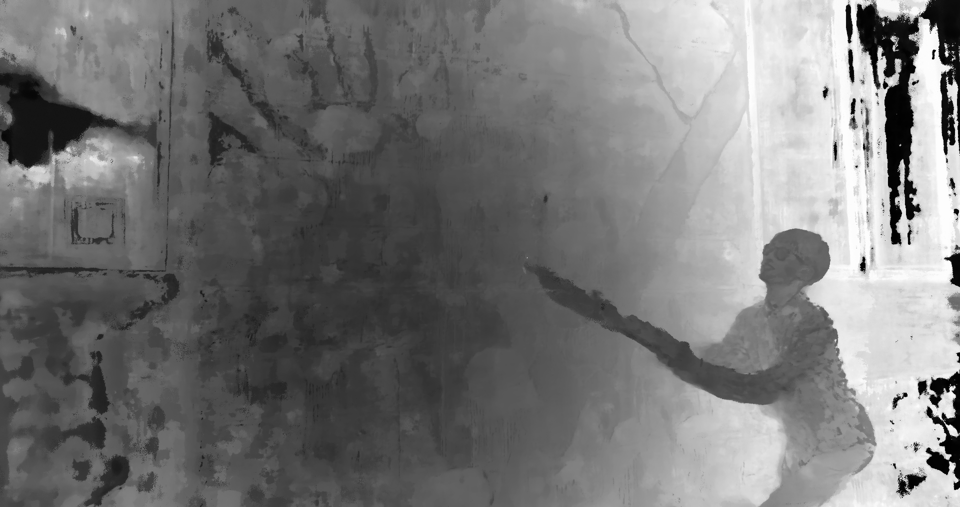}
    \end{subfigure}
\caption{4D NeRF Inpainting example. Text prompt: ``a golden sword, side view''. The first column corresponds to the first frame which includes the first seed image, and the other columns correspond to 2 later frames. Inpaint4DNeRF can generate a moving object that is overall consistent.}
\label{fig:dynamic}
\end{figure*}
\vspace{0.5mm}
\noindent \textbf{4D Example.}\quad
We show a 4D inpainting example in figure~\ref{fig:dynamic} to demonstrate that our method has the potential to generalize to dynamic NeRFs. In this example, we remove the foreground object in the video of the seed view using E2FGVI~\cite{li2022towards}, a flow-based method with optimization by feature propagation and content hallucination. For transferring motion to the generated object, after key point tracking, we estimate a rigid transformation between the key points, and propagate the pixels along the transformation. This dynamic scene consists of 16 frames, in which the first frame includes the first seed image. As shown in the figures, we successfully obtained an overall convergence on the generated object with correct motion for all the illustrated frames.

\vspace{0.5mm}
\subsection{Ablation and comparison}

To demonstrate the efficacy of various elements in our baseline design, we compare our baseline to the following 3 variants. As we are the first, to the best of our knowledge, to aim at generative NeRF inpainting, there does not exist baselines for direct comparison. Therefore, we conduct ablation by modifying some parts in our baseline, and conduct comparisons by replacing a subprocess by an existing baseline.

\begin{figure}[tp]
    \begin{subfigure}[b]{0.22\textwidth}
         \centering
         \includegraphics[width=\textwidth]{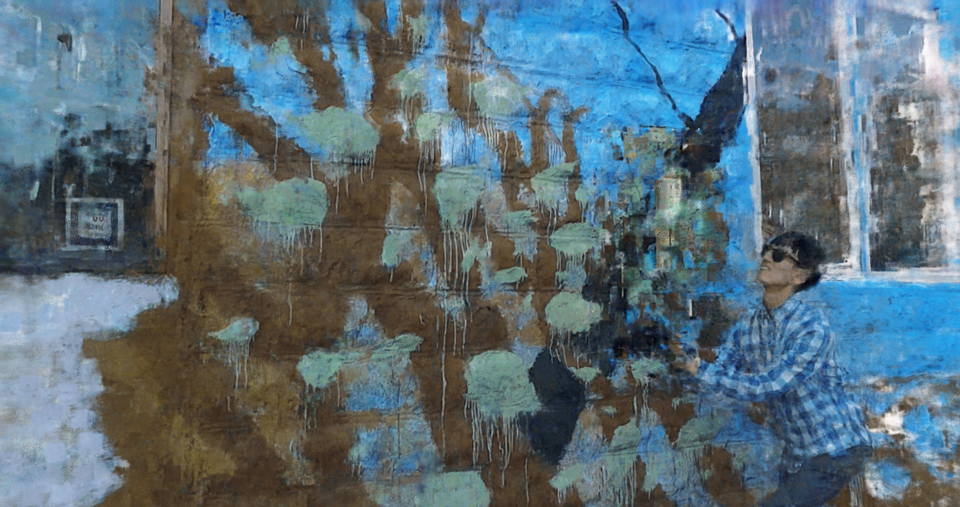}
    \end{subfigure}
    \hfill
    \begin{subfigure}[b]{0.22\textwidth}
         \centering
         \includegraphics[width=\textwidth]{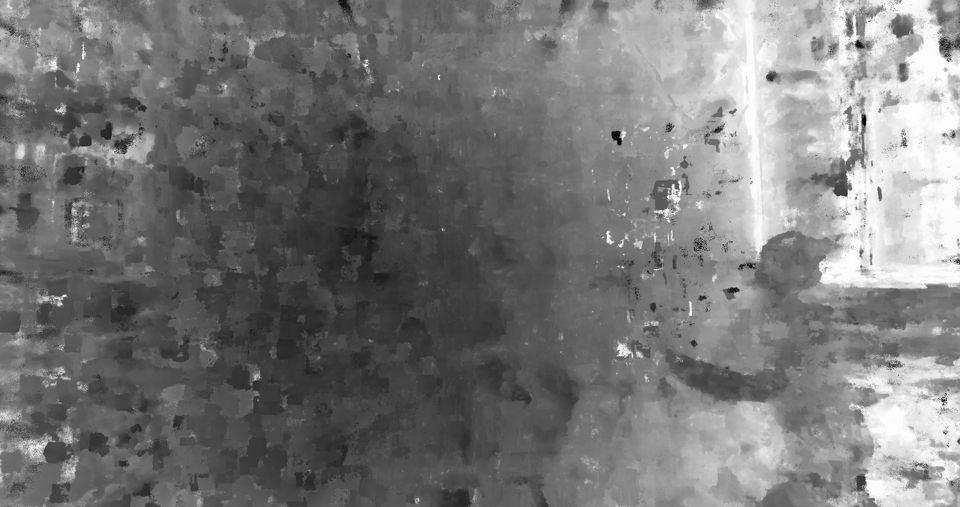}
    \end{subfigure}
    \caption{Training results with view independent inpainting. Left: rgb render. Right: noisy and incorrect depth map.}
    \label{fig:indep}
\end{figure}
\vspace{0.5mm}
\noindent \textbf{View independent inpainting.}\quad
In our baseline, all training images are pre-processed to be consistent with the first seed image, providing a guarantee for coarse convergence. To show the necessity of our pre-processing strategy, in this baseline, we inpaint all training images independently. Warmup training and IDU are kept unchanged. The results are shown in figure~\ref{fig:indep}. It is clear that independent inpainting results in divergence across views, and IDU fails to produce the initial highly inconsistent image convergence. This explains that the pre-processing step to help convergence is indispensable.

\begin{figure}[tp]
    \begin{subfigure}[b]{0.22\textwidth}
         \centering
         \includegraphics[width=\textwidth]{figs/qual/soccershirt/final_rgb.png}
    \end{subfigure}
    \hfill
    \begin{subfigure}[b]{0.22\textwidth}
         \centering
         \includegraphics[width=\textwidth]{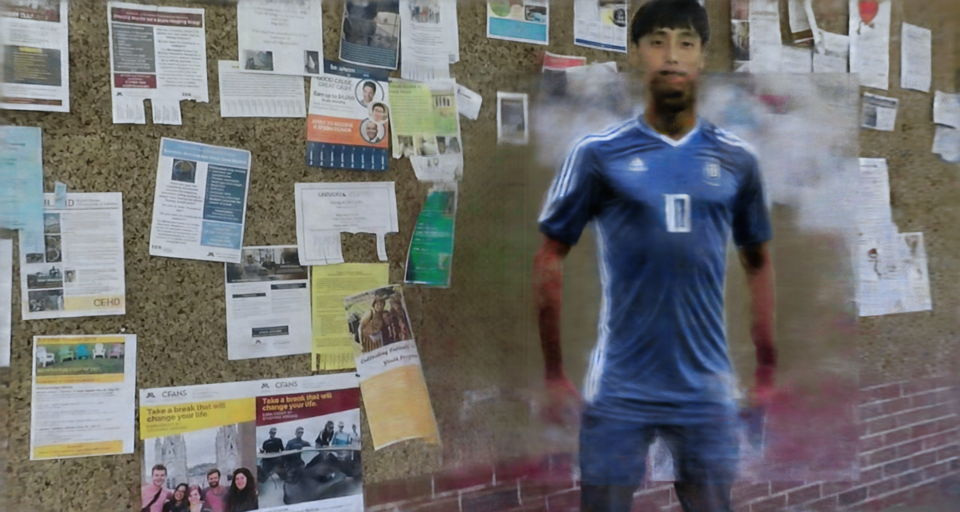}
    \end{subfigure}
    \caption{Training results with instruct-nerf2nerf. Left: Result from our baseline. Right: result from warmup training followed by instruct-nerf2nerf.}
    \label{fig:in2n}
\end{figure}
\vspace{0.5mm}
\noindent \textbf{Instruct-Nerf2Nerf after warmup.}\quad
The second stage of our NeRF training is largely based on iterative dataset update, which is the core of instruct-nerf2nerf~\cite{haque2023instruct}. This brings out the question of whether instruct-nerf2nerf can accomplish our task. Since instruct-nerf2nerf is mainly for appearance editing without large geometry changes, it is inappropriate to directly run it on our generative inpainting task. Therefore, before running their baseline, we first run our baseline until the end of warmup training, so that we have a coarse geometry as the prerequisite. By this design, the main difference between the two methods is that instruct-nerf2nerf uses instruct-pix2pix, a non-latent diffusion model, to do training image correction. Figure~\ref{fig:in2n} shows its results compared with ours. We can see that the final result maintains the blurriness from the warmup NeRF, and does not achieve a higher level of consistency.

\begin{figure}[tp]
    \begin{subfigure}[b]{0.22\textwidth}
         \centering
         \includegraphics[width=\textwidth]{figs/qual/oldprof/warmup_rgb.png}
    \end{subfigure}
    \hfill
    \begin{subfigure}[b]{0.22\textwidth}
         \centering
         \includegraphics[width=\textwidth]{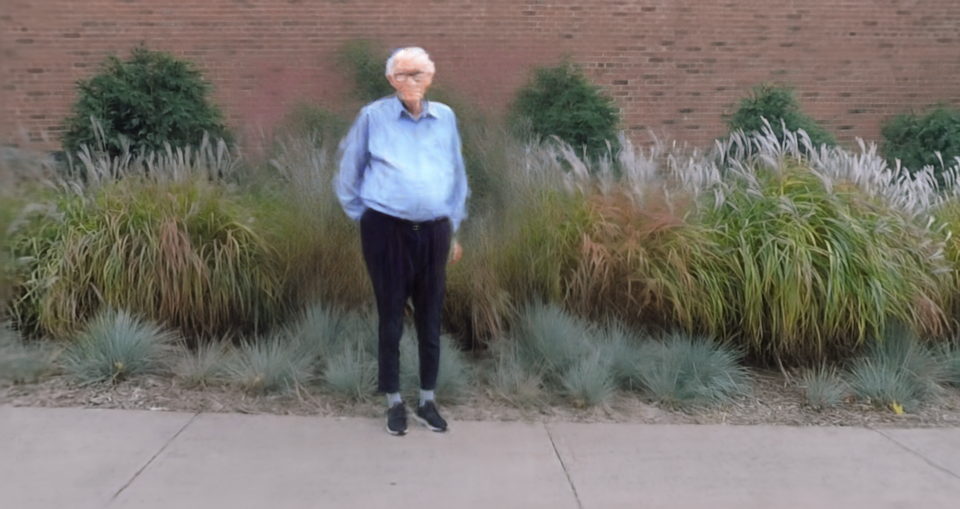}
    \end{subfigure}
    \hfill
    \begin{subfigure}[b]{0.22\textwidth}
         \centering
         \includegraphics[width=\textwidth]{figs/qual/oldprof/final_depth.png}
    \end{subfigure}
    \hfill
    \begin{subfigure}[b]{0.22\textwidth}
         \centering
         \includegraphics[width=\textwidth]{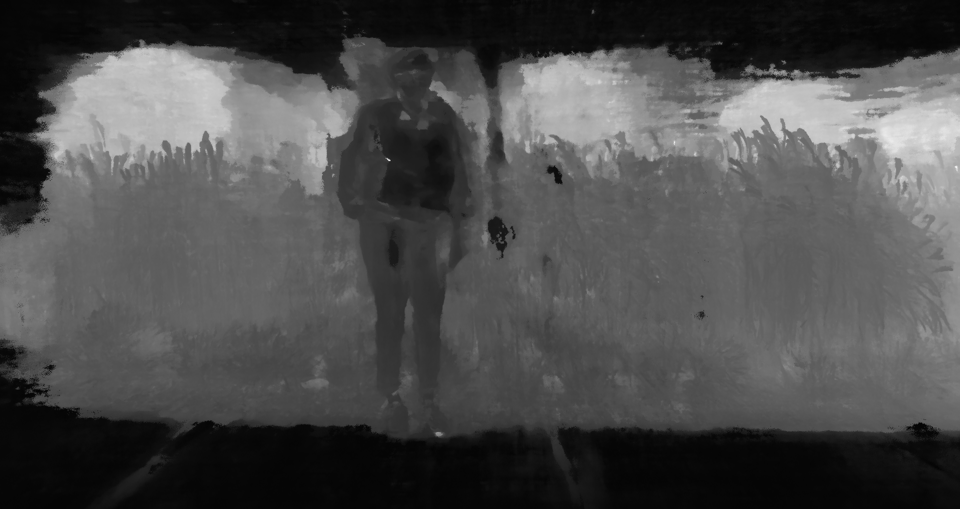}
    \end{subfigure}
    \caption{Depth maps and rgb renderings from warmup training with and without depth supervision. Left: with depth supervision (ours). Right: without depth supervision.}
    \label{fig:nodepth}
\end{figure}
\vspace{0.5mm}
\noindent \textbf{Warmup without depth supervision.}\quad
The goal of warmup training is to provide an initial convergence, especially in geometry, which requires depth loss as supervision. To demonstrate this, we provide an example of warmup training without any depth supervision, and show the rendered depth map in figure~\ref{fig:nodepth}. Compared with the depth map in our standard setting, this depth map has clear errors within and around the target object. From inconsistencies in geometry, the later fine training stage may suffer from ``floater'' and inconsistency artifacts. This can be eliminated by our simple planar depth map as guidance, even if it is not accurate.

\section{Conclusion}

We introduce Inpaint4DNeRF, a unified framework that can directly generate text-guided, background-appropriate, and multi-view consistent content within an existing NeRF.
To ensure convergence from the original object to a completely different object, we propose a training image pre-processing method that projects from initially inpainted seed images to other views, with details refined by stable diffusion. A roughly multiview consistent set of training images, combined with depth regularization, guarantees coarse convergence on geometry and appearance. Finally, the coarse NeRF is fine-tuned by iterative dataset update with stable diffusion. Our baseline can be readily extended to dynamic NeRF inpainting by generalizing the seed-image-to-other strategy from the spatial domain to the temporal domain. We provide 3D and 4D examples to demonstrate the effectiveness of our method. We also investigate the role of various elements in our baseline by ablation and comparison.

The proposed framework expands the possibilities for realistic and coherent scene editing in 3D and 4D settings. However, our current baseline still has some limitations, providing room for further improvement. Specifically, it is challenging for our method to handle complex geometry generation with a camera set covering wide angles. The consistency of the final NeRF can still be improved. In addition, to extend our method fully into 4D, certain techniques are required to further improve temporal consistency and maintain better multiview consistency across frames. We hope that our proposed baseline can inspire these future research directions for text-guided generative NeRF inpainting.
\label{sec:conclusion}

{\small
\bibliographystyle{ieeenat_fullname}
\bibliography{11_references}
}

\end{document}


\title{\paperTitle}
\author{\authorBlock}
\maketitlesupplementary

\section{Appendix Section}
Supplementary material goes here.

{\small
\bibliographystyle{ieee_fullname}
\bibliography{11_references}
}